\documentclass{article}
\usepackage{graphicx}
\usepackage{wrapfig}
\usepackage{microtype}
\usepackage{xparse,xspace}
\usepackage{amsmath,amssymb}
\usepackage{enumitem}
\usepackage{capt-of}
\usepackage{array} 
\usepackage{float}
\usepackage{tabularx}         
\newcolumntype{Y}{>{\centering\arraybackslash}X}
\usepackage{booktabs}
\usepackage[table]{xcolor}
\usepackage{nicematrix}
\NiceMatrixOptions{cell-space-top-limit=1pt,cell-space-bottom-limit=1pt}
\usepackage[font=small,labelfont=small]{caption}
\usepackage{subcaption}
\makeatletter
\renewcommand\p@subtable{}
\makeatother

\definecolor{Blue050}{HTML}{EFF6FF}
\definecolor{Blue100}{HTML}{DBEAFE}
\definecolor{Blue200}{HTML}{BFDBFE}
\definecolor{Blue400}{HTML}{60A5FA}
\definecolor{Blue600}{HTML}{2563EB}

\newcommand{\ourscell}[1]{\cellcolor{Blue100}{#1}}
\newcommand{\theadfont}{\tiny}               
\newcommand{\thc}[1]{\cellcolor{Blue050}{#1}}
\newcommand{\BCOD}{\mbox{B-COD}}

\newsavebox{\LeftTbl}
\newsavebox{\RightTbl}
\newlength{\LeftTblHeight}

\newcommand{\logoinline}[2][]{%
  \raisebox{+.3\height}{\includegraphics[#1]{#2}}%
}
\newcommand{\UIUC}{\logoinline[height=1.5ex]{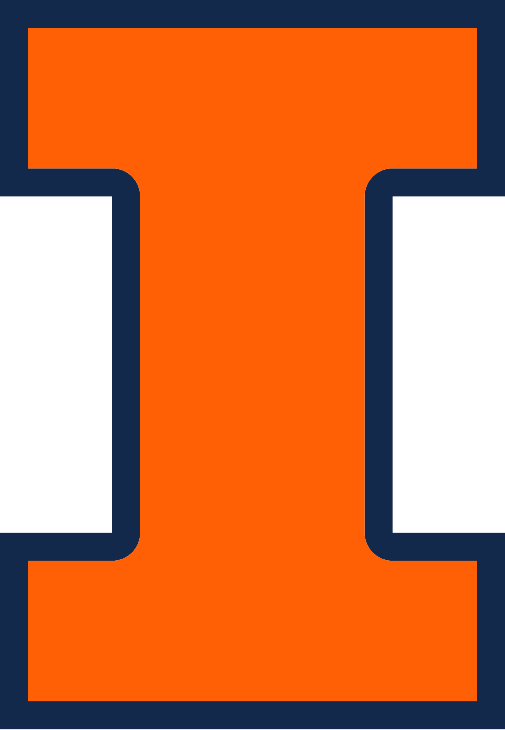}\xspace}
\newcommand{\PROV}{\logoinline[height=1.8ex]{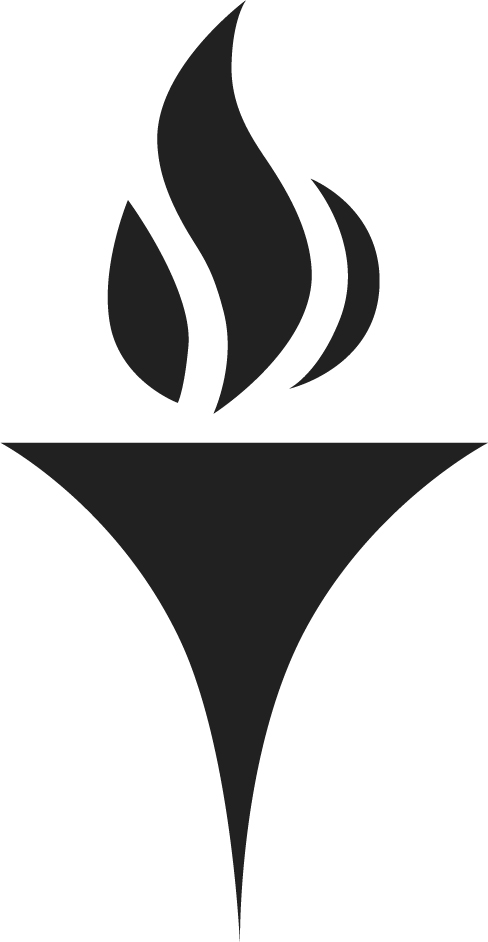}\xspace}
\newcommand{\FIU}{\logoinline[height=1.15ex]{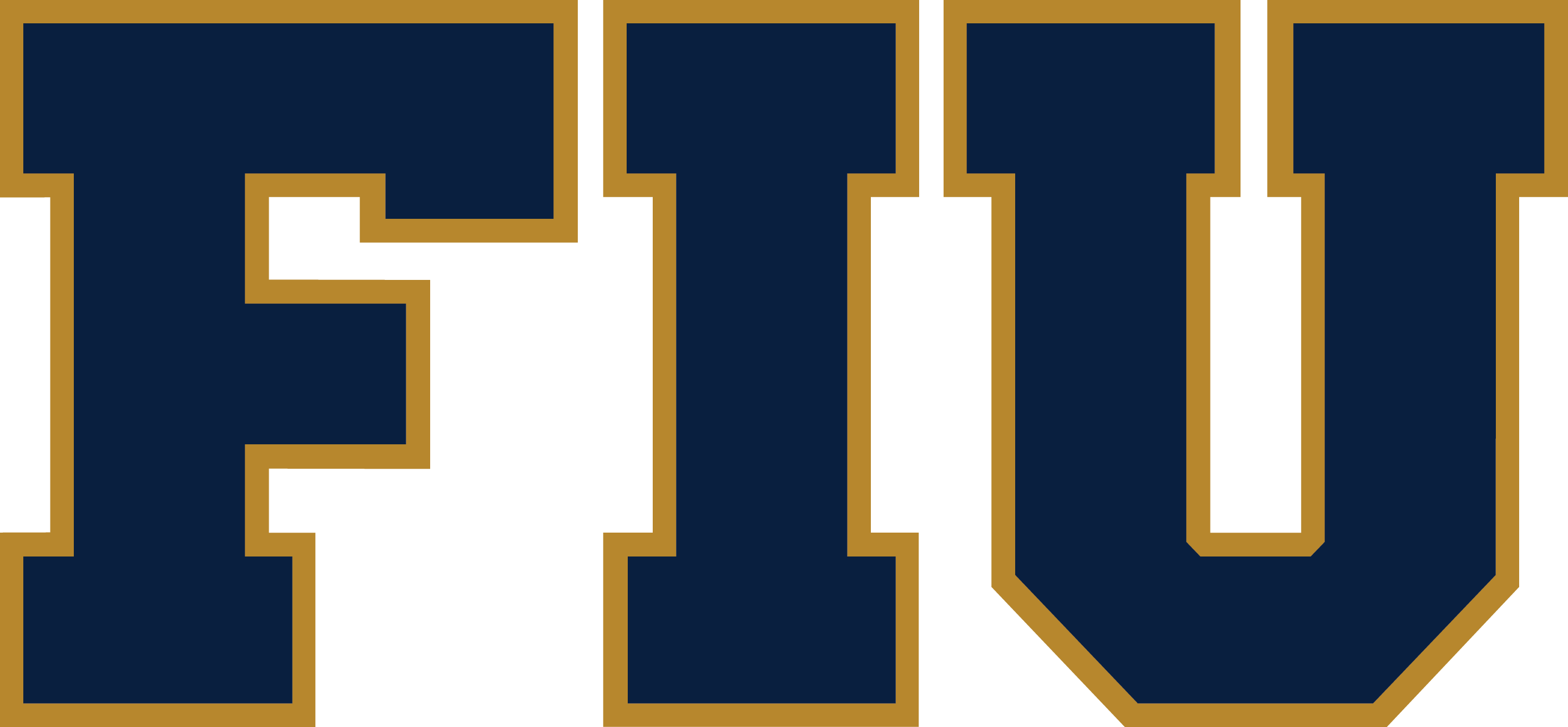}\xspace}
\newcommand{\UF}{\logoinline[height=1.3ex]{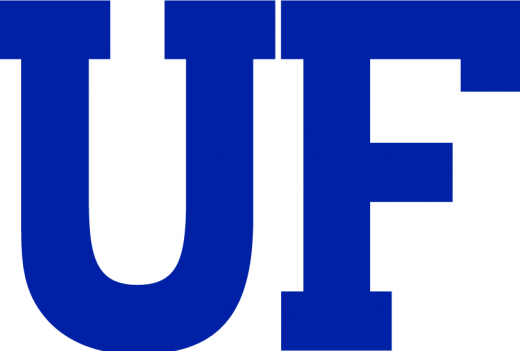}\xspace}

\newcommand{\baseline}[1]{\textit{\textcolor{black!60}{#1}}}

\newcommand{\kf}[1]{{\textcolor{Blue600}{#1}}}

\usepackage[final]{corl_2025}

\makeatletter
\renewcommand\p@subtable{}
\makeatother

\DeclareCaptionLabelFormat{subtablab}{Table~#2}
\title{Belief-Conditioned One-Step Diffusion: Real-Time Trajectory Planning with Just-Enough Sensing}

\author{
Gokul Puthumanaillam \UIUC, \quad
Aditya Penumarti \UF, \quad
Manav Vora \UIUC, \\
\textbf{Paulo Padrao}\PROV, \quad
\textbf{Jose Fuentes} \FIU, \\
\textbf{Leonardo Bobadilla} \FIU, \quad
\textbf{Jane Shin} \UF, \quad
\textbf{Melkior Ornik \UIUC} \\
\vspace{0.1cm}
\footnotesize \UIUC University of Illinois Urbana\hbox{-}Champaign (\texttt{\{gokulp2, mkvora2, mornik\}@illinois.edu}) \\
\footnotesize \UF University of Florida (\texttt{\{apenumarti, jane.shin\}@ufl.edu}) \\
\footnotesize \PROV Providence College (\texttt{\{ppadraol\}@providence.edu}) \\
\footnotesize \FIU Florida International University (\texttt{\{jfuen099\}@fiu.edu, \{bobadilla\}@cs.fiu.edu})\;}

\begin{document}
\maketitle

\begin{abstract} Robots equipped with rich sensor suites can localize reliably in partially-observable environments, but powering every sensor continuously is wasteful and often infeasible. Belief-space planners address this by propagating pose-belief covariance through analytic models and switching sensors heuristically--a brittle, runtime-expensive approach.
Data-driven approaches--including diffusion models--learn multi-modal trajectories from demonstrations, but presuppose an accurate, always-on state estimate.
We address the largely open problem: for a given task in a mapped environment, which \textit{minimal sensor subset} must be active at each location to maintain state uncertainty \textit{just low enough} to complete the task? 
Our key insight is that when a diffusion planner is explicitly conditioned on a pose-belief raster and a sensor mask, the spread of its denoising trajectories yields a calibrated, differentiable proxy for the expected localisation error.
Building on this insight, we present Belief-Conditioned One-Step Diffusion (B-COD), the first planner that, in a 10 ms forward pass, returns a short-horizon trajectory, per-waypoint aleatoric variances, and a proxy for localisation error--eliminating external covariance rollouts.
We show that this single proxy suffices for a soft-actor–critic to choose sensors online, optimising energy while bounding pose-covariance growth.
We deploy B-COD in real-time marine trials on an unmanned surface vehicle and show that it reduces sensing energy consumption while matching the goal-reach performance of an always-on baseline. 
Project website: \href{https://bcod-diffusion.github.io}{bcod-diffusion.github.io}.

\end{abstract}



\section{Introduction}
Autonomous robots performing navigation-related tasks routinely mount heterogeneous sensors, like cameras, LiDARs, and GPS, because no single modality is reliable everywhere \cite{cadena2016past, debeunne2020review, wang2020multisensor}. 
Keeping every sensor powered, however, wastes energy, and can degrade performance by flooding the estimator with irrelevant data \cite{majumdar2023fundamental, malawade2022ecofusion}.
 At the other extreme, indiscriminately toggling sensors is perilous: if the robot drifts into a sensor-denied zone with the wrong sensors active, localisation uncertainty can explode, leading to task failure \cite{kim2018selfdiagnosis, aqel2016vo_review}. 
 This paper, therefore, poses a precise question: given a goal, can a robot generate the next short-horizon motion \emph{and}, in real-time, decide the \emph{smallest} set of sensors that must stay on so that its localisation uncertainty is reduced \textit{just enough} to reach the goal---navigation with just-enough sensing.

Classical belief-space pipelines address the motion planning half of the problem: linearized covariance propagation achieves excellent tracking when every sensor is active and the motion–measurement models remain accurate \cite{rw2, berg2010lqgmp}. Yet these same pipelines unravel in harder conditions \cite{BarbosaRiskAware}, where a single ill-chosen sensor subset can drive the pose estimate outside tolerance, causing subsequent mission failure \cite{papachristos2017autonomous}.
The complementary half--deciding which sensors to power--is often posed as a scheduling problem that assigns an energy cost to every measurement \cite{ondruska2015scheduled}. The robot's belief over pose and map is continuous and high-dimensional, so the decision tree grows exponentially and is impossible to traverse in real time \cite{kaelbling1998planning}. Practical solvers collapse the tree into ad-hoc heuristics \cite{spaan2009dynamic}, thereby ignoring the long-term value of information \cite{charrow2015information}.
Reinforcement learning variants push farther by embedding the sensor-activation bits into the action space and letting a policy learn when to switch them \cite{leong2020deep, alali2024drlschedule, comtraqmpc}. However, such policies demand millions of episodes, delicate reward tuning \cite{akkaya2019solving}, and still require hand-engineered safety shields to keep pose error bounded \cite{alshiekh2018safe}.  
Worse, they optimize sensing and control on the same lattice but leave trajectory generation to a separate planner, creating a brittle split that resurfaces whenever the environment changes \cite{malawade2022ecofusion, ondruska2015scheduled, honda2024whentoreplan}.

\begin{figure}
    \centering
    \includegraphics[width=0.8\linewidth, trim=20 250 35 25,clip]{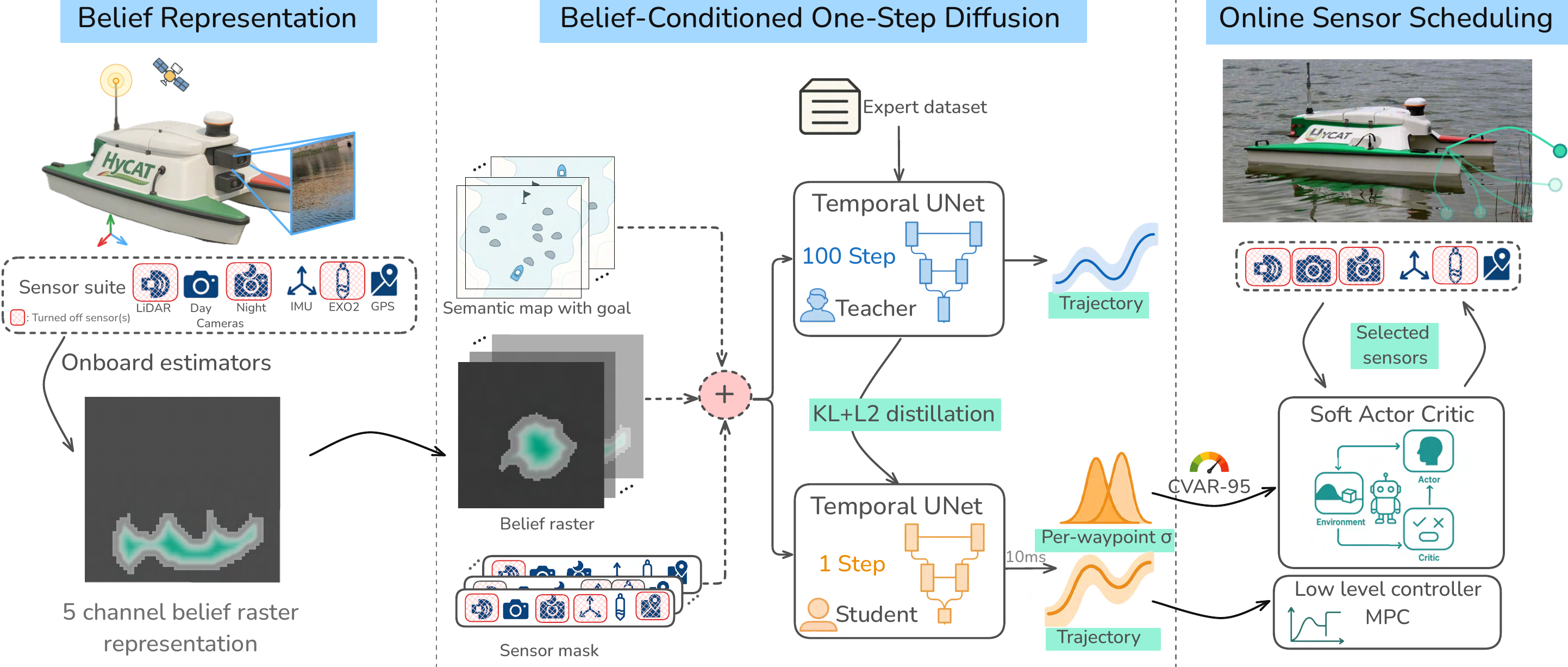}
    \caption{Overview of B-COD. Left: The belief module compresses the particle cloud and local map into a belief raster that encodes mass, orientation, and covariance.
Center: A one-step diffusion network consumes that raster, the goal mask, and the current sensor flag to return a short-horizon trajectory and a calibrated CVaR risk scalar.
Right: SAC uses the scalar to toggle sensors in real time, spending energy only where required.}
    \label{fig:BCODRep}
\end{figure}

In a departure from handcrafted motion models and data-hungry on-policy RL, diffusion-based planners learn rich, multi-modal trajectory distributions directly from demonstrations and require only supervised training \cite{janner2022diffuser, luo2024potential, chi2023diffusionpolicy}. Their appeal is clear: they avoid reward-shaping pitfalls, capture alternate homotopies around obstacles, and can be conditioned on semantic maps with minimal architectural fuss. Their Achilles' heel, however, is that the denoising network is trained under the premise that the robot's pose is always known accurately and every sensor is live \cite{carvalho2023mpdiffusion, sun2023plancp}. 
A largely overlooked aspect of diffusion planners is that their iterative denoising process \cite{ho2020denoising} produces a \textit{full distribution} of trajectories rather than a single path \cite{janner2022diffuser, feng2024tat}. Intuitively, the spread of that distribution ought to widen when localisation drifts and contract when well observed–-hinting
that diffusion \textit{might} already encode clues about ``how much sensing is enough." Turning that intuition
to enable navigation with just-enough sensing is the central idea of this paper: we repurpose diffusion to simultaneously produces a short-horizon path towards the goal \textit{and} serve as a sensing oracle for a small sensor-selection policy. Figure~\ref{fig:BCODRep} presents an overview of our solution approach.

\textit{Statement of contributions:}
We introduce (i) \textit{Belief-Conditioned One-Step Diffusion (B-COD)}, a diffusion planner  that pairs short-horizon trajectory generation with an uncertainty proxy from its own denoising spread, generated in one forward pass; (ii) we show that this proxy is enough to enable a lightweight RL policy to toggle sensors online, yielding a real-time navigation pipeline with just enough sensing; (iii) we validate the full system in real-time on an autonomous surface vehicle in the context of marine autonomy, an underexplored, high-uncertainty domain, demonstrating goal-reach rate of the always-on baseline while consuming less than half its energy; 
(iii-a) to spur further research in this domain, we release the code, trained models, and the first open maritime dataset of 50K belief-annotated navigation snippets with synchronised multi-modal sensor logs
\footnote{A subset is available on the project website; the full dataset will be released after review.}.



\section{Related Works}
\textit{Belief-Space Motion Planning: }Early work cast motion planning under uncertainty as a POMDP, but exact solvers scale poorly, motivating approximate methods \cite{rw1, rw2, rw3}. 
Graph-based variants propagate linearised covariances along lattice edges \cite{a1, a2, a3, a4, a5} or sample belief particles through non-linear dynamics \cite{a6, a7, a8, a9}. 
Point-based POMDP planners \cite{b1, b2, b3, b4, b5, b6} achieved impressive accuracy but remain slow for real-time use. Chance-constrained formulations bound the probability of constraint violation \cite{b7, b8, b9, b10}, while information-theoretic objectives \cite{c1, c2, c3, c4} trade control cost with covariance growth. These pipelines assume all sensors remain active and rely on expensive covariance roll-outs at every node--limitations B-COD sidesteps by learning a differentiable proxy in one forward pass.

\textit{Resource-Aware Sensor Scheduling: }Choosing which modalities to power has been framed as a mixed-integer program over measurement actions \cite{d1, d2, d3, d4}, a sub-modular maximisation of expected information gain \cite{d5, d6, d7}, or a value-of-information heuristic \cite{e1, e2, e4}. Sensor-activation trees grow with horizon, so practical solvers rely on greedy switches \cite{e5, e6}, rollout policy iteration \cite{e7, e8}, or coarse duty-cycling \cite{e9, e10}. RL approaches embed sensor bits into the action space and learn policies offline \cite{f1, f2, f3, f4, f5, f6}. Energy-aware schemes \cite{f9, f8, f9} toggle sensors based on heuristic. Unlike these decoupled strategies, B-COD couples trajectory generation and sensor selection via a shared diffusion-derived uncertainty signal without handcrafted thresholds.

\textit{Learning-Based Trajectory Generators: }Imitation-learning priors \cite{g1, g2, g3, g4, g5} capture multi-modal behaviour but output a single trajectory. VAEs \cite{g5, g6, g7, g8, g9} model richer path distributions, yet training requires tricks or annealing. Diffusion-based planners \cite{h1, h2, janner2022diffuser, chi2023diffusionpolicy, rw7, bouvier2025ddat} have recently shown SOTA coverage over homotopies while retaining simple supervised losses. All, however, assume perfect, always-on estimation. B-COD is the first to inject the robot's pose-belief raster into the denoiser, letting the spread of the samples act as a calibrated localisation-error proxy.

\textit{Joint Planning and Active Perception:} Active SLAM couples view-planning with mapping updates but focuses on map quality rather than energy \cite{h4, h5, h3}. Informative path planners \cite{h6, h7} optimise mutual information over candidate trajectories while assuming fixed sensor payloads. Recent works \cite{h8, h9} integrate learned priors with active perception yet they still separate motion generation from modality selection, requiring auxiliary covariance estimators. By conditioning diffusion on belief and letting an RL head act on the resulting uncertainty proxy, our framework realises just-enough sensing--closing the loop between planning and active localisation with millisecond latency.

\section{Methodology}
\textbf{Problem statement. }Assume a robot operating in a workspace $\mathcal W\subset\mathbb R^{2}$ whose obstacles and semantic layers--lighting, sensor visibility--are known in advance through survey or GIS data, yielding a coarse, static map \(M\). The robot carries $N$ sensors $S=\{s_1,\dots,s_N\}$. At decision epoch \(t\) the robot's pose is the element $\mathbf{x}_t=(x_t,y_t,\psi_t)\in SE(2)$; its uncertainty is encoded by the belief density $b_t(\mathbf{x})=p\bigl(\mathbf{x}_t=\mathbf{x}\mid z_{0:t},a_{0:t}\bigr)$,
where $z_{0:t}$ are past measurements and $a_{0:t}$ is the sensor-activation vectors. Observations obey a known likelihood $p(z_t\mid\mathbf{x}_t,a_t)$. At each step, $a_t\in\{0,1\}^N$ indicates which sensors are powered, and the energy cost vector $c=[c_1,\dots,c_N]^\top$ quantifies per-sensor consumption. Together with a control sequence \(\tau_t\), the schedule \(\{a_t\}_{t=0}^{T-1}\) must steer the belief into the goal set \(\mathcal G\) while avoiding the obstacle region \(\mathcal O\).
We measure localisation risk by the scalar $\sigma(b_t)=\sqrt{\operatorname{tr}\,\Sigma_t}$
where \(\Sigma_t\) is the pose covariance extracted from $b_t$. Let $\eta$ denote the user-specified per-step risk budget and $\varepsilon$ the tolerated per-step exceedance probability. The planning problem is\nobreak
\begin{equation}
\text{minimize } 
\;J_{\text{energy}}=\sum_{t=0}^{T-1} c^{\top} a_t
\quad\text{subject to}\quad
\Pr\!\bigl\{\sigma(b_t)>\eta\bigr\}\le\varepsilon,\;
\mathbf{x}_t\notin\mathcal O,\;
\mathbf{x}_T\in\mathcal G,\;\forall t,
\end{equation}

Our solution hinges on three capabilities: (i) a belief encoding that neural networks (NN) can ingest; (ii) a planner that, conditioned on that belief, yields a trajectory and a measure of future pose uncertainty; (iii) a policy that exploits that measure to enable minimal sensing task completion.  

\subsection{Belief Raster Representation}\label{bel}
\begin{wrapfigure}[18]{r}{0.35\textwidth} 
  \vspace{-6pt}           
  \centering
  \includegraphics[width=\linewidth]{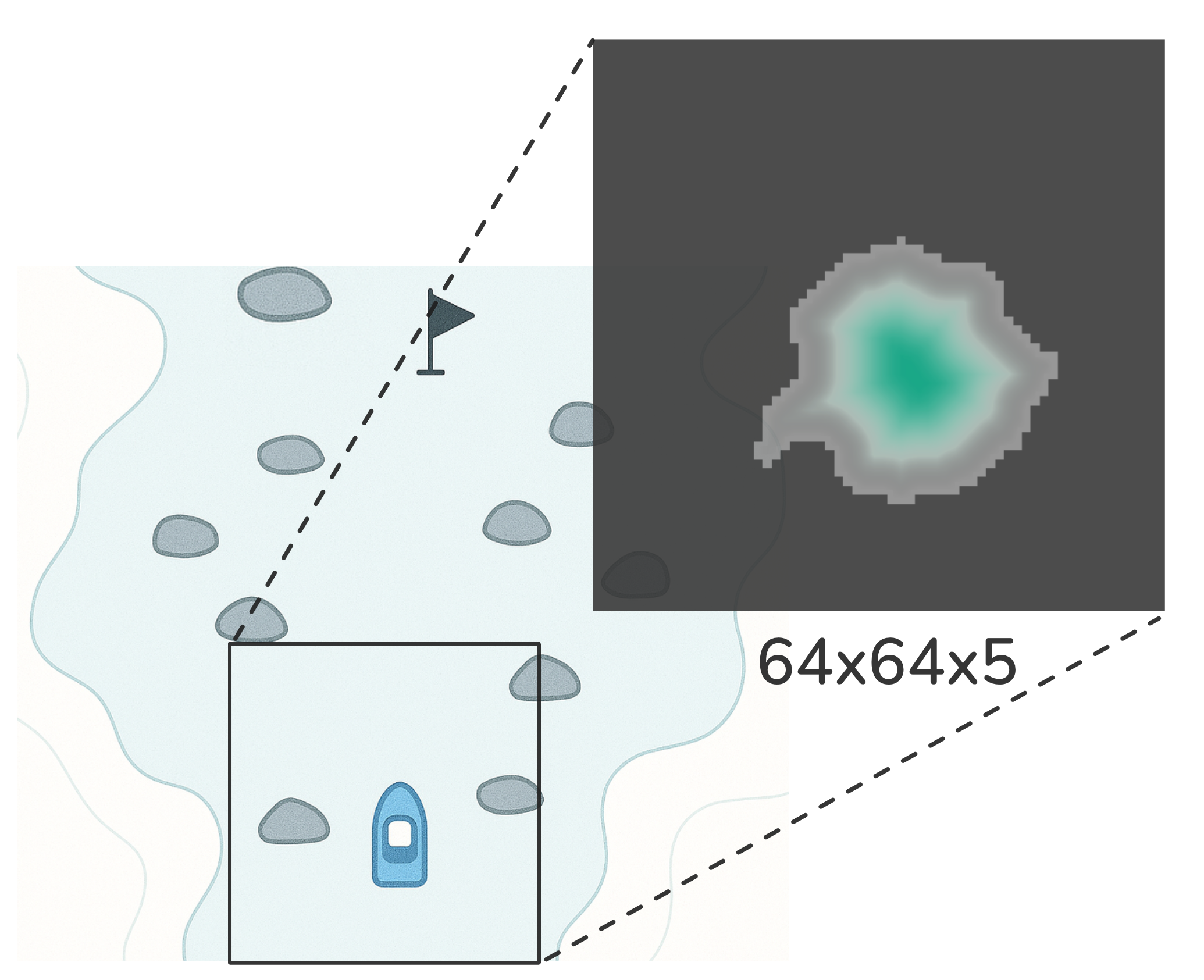}
  \caption{Example belief raster from our experiments.
The channels are projected into a single HSV image: brightness = probability mass, hue = heading, desaturation = areas of higher positional spread. The teal = most likely pose; the grey expresses growing 3$\sigma$ uncertainty, black denotes zero belief.}
  \label{fig:arm-right}
\end{wrapfigure}
Any NN planner that consumes state uncertainty must see a tensor of fixed shape \cite{nguyenmotionprim2022}, yet a Bayesian filter evolves a belief whose support may swell/split over time \cite{dellaert_monte_1999}.  
We therefore project the pose posterior into a five-channel image $B_t\in[0,1]^{H\times W\times 5}$ whose footprint adapts to the belief but is resampled to a preset resolution before it enters the planner.
If the window exceeds the target lattice $H\times W$ it is isotropically down-sampled, making runtime independent of how far the robot's uncertainty has spread. 
Each grid cell stores five summary statistics that experiments show are sufficient:
(i) \textit{belief mass} \(m_{u,v}\), a true probability derived by summing particle weights in the cell, 
(ii-iii) \textit{sine} and \textit{cosine of yaw} that encodes the heading, averaged with the same weights and linearly re-mapped to \([0,1]\) so that all channels share a common dynamic range,  
(iv) \textit{planar spread} is compressed into the log-determinant of the local positional covariance,
(v) \textit{circular variance} records how concentrated the heading distribution is; it equals zero for a unimodal orientation and approaches one as yaw becomes ambiguous.
Other moments were found to add negligible value (see Appendix).  
$B_t$ is translation-equivariant, orientation-aware, and compact enough to slot directly into downstream convolutions, yet retaining exactly the information that governs localisation drift for small horizons.


\subsection{Belief-Conditioned One-Step Diffusion (B-COD)}\label{bcod}
Having condensed pose uncertainty into an ego-centric image, we face two coupled decisions at every control tick: what short motion should the robot execute next, and how large will the pose error grow if it moves with only the currently powered sensors?
B-COD answers this by learning the conditional distribution
$p_\theta\!\bigl(\tau \,\big|\, B,\;M,\;g,\;a\bigr),$
where \(B\) is the ego-centric belief image from Sec.~\ref{bel}, \(M\) the co-cropped semantic map slice, \(g\) a binary goal mask in the same frame, \(a\in\{0,1\}^N\) the vector of powered sensors, and  
\(\tau=(\Delta x_1,\Delta y_1,\Delta\psi_1,\dots,\Delta\psi_{H})\) 
sequence of robot body-frame incremental poses.
We noticed that using increments, rather than global coordinates, lets B-COD learn translation-invariant manoeuvres; as B-COD is re-invoked each second, these local segments concatenate naturally into a consistent global path.
The diffusion backbone parameterises, at every waypoint, a diagonal Gaussian whose mean is the desired displacement and whose log-variance $\hat\sigma_k$ quantifies aleatoric uncertainty inherited from the training data.
Drawing one standard normal latent and passing it through the network therefore returns in a \emph{single} forward pass (i) a trajectory sample \(\tau\) and (ii) the full vector of waypoint log-variances \(\hat\sigma_{1:H}\).
Resampling the latent injects fresh noise and produces alternative, equally plausible plans without any extra cost beyond that single inference--exactly the property that makes diffusion attractive for real-time, multi-modal navigation.

\textbf{Conditioning and trajectory tokenisation.} To generate a path that is feasible and sensor-aware, the planner must jointly reason over: (i) \emph{where the robot might be}, (ii) \emph{what the environment looks like}, and (iii) \emph{which modalities are currently online}.
The inputs that carry this context--belief, map semantics, and goal--are concatenated into a tensor \([B\,\|\,M\,\|\,g]\). 
Since navigation occurs locally, encoding the entire global map is redundant and inefficient; thus, a generic spatial encoder (in our case, ResNet) can extract spatially localized features that summarise nearby geometry and uncertainty \cite{zhao_review_2024}.
Additionally, because trajectories depend on the available sensors, an embedding of \(a\) is fused with the spatial features to form a comprehensive context vector \(\mathcal{C}\). 
Conditioning on $\mathcal{C}$ ensures that the planner avoids plans that would demand unavailable modalities.
Finally, the planner outputs trajectories as sequences of relative displacements rather than global poses. This representation makes the prediction invariant to absolute coordinates, allowing the diffusion network to attend directly to the local environment and sensor context vector $\mathcal{C}$. 
This conditioning dynamically aligns each waypoint with obstacles, environmental context, and real-time sensing capabilities. 
\textbf{Diffusion teacher.}  
A diffusion model can sample diverse trajectories, but running a hundred reverse steps is prohibitive on embedded hardware \cite{h1}.
We therefore follow the DDPM paradigm \cite{ho2020denoising}: teach a multi-step denoiser that is \emph{expressive}--able to regenerate the multi-modal paths seen in demonstration--and \emph{honest} about its own pose uncertainty. 
The teacher adopts the cosine forward process \cite{Chen2024AnOO}, \(\tilde\tau_t=\sqrt{\bar\alpha_t}\,\tau+\sqrt{1-\bar\alpha_t}\,\epsilon\), \(\epsilon\!\sim\!\mathcal N(0,I)\) with \(\bar\alpha_t=\cos^2(\pi t/2T)\), because that schedule keeps signal-to-noise high at early timesteps and has been shown to improve reconstruction quality on long, structured sequences \cite{Chen2024AnOO}. 
At reversal step \(t\) the network predicts both the injected noise and a waypoint-wise log-variance--crucial, because how confident the network is about each displacement becomes the proxy our scheduler relies on.
The loss combines the DDPM noise-reconstruction term with a diagonal Gaussian negative log-likelihood:
\begin{equation} \label{teacher}
\mathcal L_{\text{teach}}=
\bigl\lVert \epsilon_\theta(\tilde{\tau}_t,B,M,g,a,t)-\epsilon \bigr\rVert_2^{2}
+\beta\sum_{k=1}^{H}\Bigl[
      \tfrac{\lVert\tau_k-\hat\mu_{\theta,k}\rVert^{2}}{e^{\hat\sigma_{\theta,k}}}
      +\hat\sigma_{\theta,k}
   \Bigr].    
\end{equation}
The first term teaches the network to undo the forward corruption and reproduce the expert manoeuvre; the second forces the predicted variance \(e^{\hat\sigma_k}\) to match the empirical scatter of demonstrations at that waypoint, yielding aleatoric errors that are \textit{calibrated by construction}.  

\textbf{Consistency-model student.}  
Even a perfectly trained teacher is unusable if their hundred reverse steps exceed the control loop's latency.  Rather than prune steps heuristically--which degrades sample quality\cite{Fan2025ASO}--we compress the reverse chain into a single network via consistency distillation \cite{Song2023ConsistencyM}. A standard-normal latent \(\xi\) is pushed through the teacher to obtain a reference sample \((\tau^{\text{ref}},\hat\sigma^{\text{ref}})\).  
A student \(g_\psi\) is then optimised to match both the mean path and the uncertainty of that reference:
\begin{equation}
\mathcal L_{\text{stu}}=
\bigl\lVert g_\psi^{\mu}(\xi)-\tau^{\text{ref}}\bigr\rVert_{2}^{2}
+\lambda\;
   \mathrm{KL}\!\Bigl(
     \mathcal N\!\bigl(g_\psi^{\mu},\Sigma_\psi\bigr)
     \,\big\|\,
     \mathcal N\!\bigl(\tau^{\text{ref}},\Sigma_{\text{ref}}\bigr)
   \Bigr),
\qquad
\Sigma=\operatorname{diag}\!\bigl(e^{\hat\sigma}\bigr).    
\end{equation}
The first term preserves trajectory fidelity; the KL term transfers the teacher's calibrated variances so the risk proxy remains valid after compression.
We ramp \(\lambda\) during training so that the student first learns accurate means and only then assumes responsibility for matching dispersion--preventing the early collapse of variance that can plague single-step models \cite{Fan2025ASO}.

\textbf{Uncertainty proxy for the scheduler.}  
Passing the whole variance vector to the planner would explode the state space; collapsing it to a naive average would hide the critical segments where error spikes. 
We condense the vector into a tail–focused statistic, CVaR-95 \cite{Rockafellar2000OptimizationOC}:
$u^{\text{CVaR}}
=\frac{1}{0.05\,H}\sum_{k\in\text{top }5\%}\!\sqrt{e^{\hat{\sigma}_k}}.$
Intuitively, because the teacher is trained with a diagonal Gaussian NLL and the student matches that distribution via the KL term, the CVaR bound transfers intact: on the training distribution the one-sided 95th-percentile localisation error at \emph{each} waypoint does not exceed \(u^{\text{CVaR}}\). Temporal correlations can only render the statistic conservative, a property we deliberately preserve for safety.
Crucially, the scalar drops out of the planner's existing variance head--no extra sampling or covariance rollout is required--so it arrives alongside the candidate trajectory within the same forward pass. 

\subsection{Online sensor scheduling with a risk budget}\label{sac}
At every decision epoch, B-COD delivers a short-horizon trajectory and \(u^{\text{CVaR}}\). 
The robot must now answer the question: which subset of sensors should remain powered so the error budget is honoured while energy is spent as sparingly as possible?
We cast this as a \textit{Constrained Markov Decision Process (CMDP)} \cite{altman2021constrained} with state  
\(s_t=\bigl(B_t,\;u^{\text{CVaR}}_t,\;d_t,\;a_{t-1}\bigr)\): belief raster, scalar risk forecast, distance from the belief mean to the goal, and the previously active sensor mask. 
The primary cost is instantaneous power draw \(r_t=-c^{\!\top}a_t\) with per-sensor coefficients \(c\in\mathbb R_{>0}^{N}\).  The indicator constraint expresses safety  
\(g_t=\mathbf 1[u^{\text{CVaR}}_t>\eta_{\max}]\): incurs a penalty if the predicted localisation error exceeds the defined threshold \(\eta_{\max}\). 
Letting \(\pi\) denote any stationary policy, the optimisation objective is  
\begin{equation}
\min_{\pi}\;
\limsup_{T\to\infty}\frac1T\,
\mathbb E_\pi\!\Bigl[\sum_{t=0}^{T-1} r_t\Bigr]
\quad\text{s.t.}\quad
\limsup_{T\to\infty}\frac1T\,
\mathbb E_\pi\!\Bigl[\sum_{t=0}^{T-1} g_t\Bigr]
\le \epsilon ,
\end{equation}
where \(\epsilon\) is the desired long-run rate of risk violations. We append the numerical features \((u^{\text{CVaR}}_t,d_t,a_{t-1})\) to the encoded $B_t$ from B-COD and feed the result to a lightweight policy head.
Any constrained-RL can solve this CMDP; we instantiate it with a \textit{constrained Soft Actor–Critic} (SAC) \cite{haarnoja2018soft, puthumanaillam2024enhancing} because it is off-policy, data-efficient, and admits a simple dual-gradient update on the Lagrange multiplier \(\lambda\). 
The actor outputs relaxed Bernoulli probabilities thresholded to hard on/off commands at execution time, while a pair of critic networks learns the Lagrangian Q-values. 
The only hyperparameters exposed are the risk budget \(\eta_{\max}\) and the violation rate \(\epsilon\).


\subsection{Engineering Choices and Implementation Recipe}
\textit{Data generation: } Alongside real-world data collection, we augment data from a custom Unity-based simulator that instantiates ninety procedurally varied harbour scenes ($\approx200~m^2$). Each world is rasterised at 0.25 m resolution into three semantic layers--obstacle occupancy, ambient lighting, and sensor visibility--cropped into the ego frame alongside the belief raster.  An A*+MPC \cite{4082128, garcia1989model} oracle drives from random starts to six-meter goal disks while seeing ground-truth pose. Replaying every oracle trajectory sixteen times under forced sensor subsets and noise produces 8.3M short-horizon snippets whose diversity teaches the B-COD both nominal and failure-case behaviour.

\textit{Training: }The diffusion teacher is trained once on that corpus with the Eq.~\ref{teacher} loss; an exponential-moving-average of the weights is retained as the final teacher. A single-step student is consistency distilled training on 1M noise targets generated on-the-fly, giving a forward pass two orders of magnitude faster yet matching its negative log-likelihood. Exporting the student through ONNX \cite{ONNX2019} and TensorRT \cite{NvidiaTensorRT2023} (FP16, layer fusion) yields a $\approx$ 10 ms inference on the Jetson-class hardware (Orin NX 16 GB \cite{NvidiaJetsonOrinNX2024}); belief rasterisation and actor inference raise the control-loop budget to 15 ms (empirical results in Sec.~\ref{sec:result}). SAC learns the scheduling policy entirely in simulation. A shared CNN \cite{krizhevsky2012imagenet} processes the belief image once; two MLPs serve as the lightweight policy head and output Bernoulli logits and twin Q-values. Dual-gradient updates on the Lagrange multiplier keep constraint violations near the user-specified budget.
In real-time testing, the neural checkpoints run unchanged. Only sensor-noise covariances are re-scaled from factory calibration data.  
\section{Experimental Results} \label{sec:result}
Our evaluation targets a real-world, real-time scenario in which an autonomous surface vehicle (ASV) must navigate an open-air lake previously unseen in training, to reach waypoint goals with just-enough sensing while keeping the CVaR-95 localisation error below a user budget of 2 m. 
The lake presents both natural and human-driven disturbances: winds, waves, fountains, and floating buoys.
The test platform is a SeaRobotics Surveyor ASV \cite{SeaRoboticsSurveyor2025} with a differential-thrust propulsion module and a heterogeneous sensor suite: a multi-beam LiDAR, day and night cameras, RTK-GPS, MEMS IMU, and an EXO2 sonde.
The platform exposes a velocity set-point interface; we run a linear MPC (2 s horizon; inputs limited to the thrust envelope) that tracks an eight-waypoint segment with mean cross-track error $\leq$ 6 cm. Control inputs are augmented by a discrete mode flag that selects the estimator configuration implied by the powered sensors. Sensor power draw differs by an order of magnitude, so efficient scheduling has tangible impact on total mission energy (see Appendix). 

\subsection{Baselines}
We benchmark B-COD against alternatives chosen to isolate each of our findings while sharing the same low-level controller (see Appendix for hyperparameters). 
\baseline{(i) Always-ON} keeps every sensor powered and sets the upper bound on task success. \baseline{(ii) Greedy-OFF} is a hand-tuned rule that asks whether simple heuristics obviate learning: day camera disabled below 10 lux, night camera disabled above 10 lux, LiDAR spun up if its previous sweep reported a return within 15 m, GPS + IMU always on, EXO2 sonde enabled in high-interest water. The values are grid-searched on a held-out set. 
To test whether a principled heuristic suffices, \baseline{(iii) InfoGain-Greedy} keeps every sensor off except the one predicted by the analytic observation model to maximise single-step expected entropy reduction of the Gaussian-mixture belief. 
\baseline{(iv-v) Random-K} controls sensing duty-cycle by sampling one or two sensors, irrespective of context.
To verify that the advantage stems from the risk proxy, we keep the SAC unchanged and swap its inputs: \baseline{(vi) $\sigma$-Mean} feeds the actor the mean of B-COD's waypoint standard deviations; \baseline{(vii) Sample-Spread} drops the variance head and estimates risk from the empirical spread of 20 Monte-Carlo trajectory samples. Baselines (i)-(vii) keep the B-COD planner frozen. 
To probe representation: \baseline{(viii) No-Belief Raster} trains the diffusion model on a single delta-pose channel only. 
We benchmark against alternative planners: \baseline{(ix) Pure-RL} lets a constrained-SAC \cite{9766421, puthumanaillam2024guided} learn both motion primitives and sensor toggles end-to-end and is trained for 50 M environment steps.
\baseline{(x) DESPOT-Lite} \cite{somani2013despot}, runs a 5k-particle online POMDP search with analytic observation models. 
\subsection{Key Findings}
\kf{Key Finding \#1: B-COD+SAC delivers near-perfect task completion at less than half the sensing cost of the Always-ON baseline.}
Table~\ref{tabres} summarizes performance over 50 laps. 
B-COD reaches the goal on 97.9 \% of attempts, yet spends only 42 \% of the energy.  
Collisions remain at 0.9 \%, essentially identical to the Always-ON baseline. 
Heuristic scheduling cannot match this trade-off: Greedy-OFF conserves energy (61 \%) but sacrifices success (47 \%). 
InfoGain-Greedy raises success to 90 \% yet violates risk eight times more often than B-COD.  
Random masks fare worse, proving that local environment context--not just a lower duty cycle--is essential for task completion.
Pure-RL generates trajectories and schedules sensors from raw rasters; the high-dimensional action space makes exploration sparse, and the policy converges to risk-averse dithering--only 55 \% goals reached and a 22 \% collision rate. DESPOT-Lite, by contrast, evaluates a principled belief tree with analytic models and therefore is able to plan accurately, but it expands hundreds of nodes; the resulting 0.5s runtime renders it unusable in real-time on the vehicle.

\begingroup

\begin{table*}[h!]
  \centering

  \sbox{\LeftTbl}{%
    \begin{minipage}[t]{0.69\textwidth}
      \centering
      \begingroup
        \scriptsize
        \setlength{\tabcolsep}{2pt}

        \newcolumntype{C}{>{\centering\arraybackslash}X}
        \newcolumntype{S}{>{\hsize=.9\hsize\centering\arraybackslash}X}

        \begin{tabularx}{\textwidth}{@{}l!{\vrule width 0.1pt}C S S S S S S C S C C@{}}
          \toprule
          {\theadfont\thc{Metric}} &
          {\theadfont\thc{\baseline{AON}}} &
          {\theadfont\thc{\baseline{GOF}}} &
          {\theadfont\thc{\baseline{IGG}}} &
          {\theadfont\thc{\baseline{R1}}} &
          {\theadfont\thc{\baseline{R2}}} &
          {\theadfont\thc{\baseline{$\sigma$M}}} &
          {\theadfont\thc{\baseline{SS}}} &
          {\theadfont\thc{\baseline{NB}}} &
          {\theadfont\thc{\baseline{PRL}}} &
          {\theadfont\thc{\baseline{DL}}} &
          {\theadfont\thc{\baseline{\BCOD}}} \\
          \midrule
          \theadfont Goal-reach (\%) $\uparrow$  & 100.0 & 47.3 & 89.9 & 18.5 & 29.1 & 79.6 & 94.3 & 67.8 & 54.8 & 87.9 & \ourscell{97.9} \\
          \theadfont Collision (\%)  $\downarrow$ & 0.5 & 22.3 & 6.1 & 34.5 & 30.1 & 12.4 & 4.7 & 17.4 & 22.1 & 4.2 & \ourscell{0.9} \\
          \theadfont CVaR violations (\%)  $\downarrow$ & 0.1 & 15.8 & 4.3 & 28.6 & 22.8 & 9.1 & 5.2 & 13.2 & 18.3 & 1.9 & \ourscell{0.5} \\
          \theadfont Mean \#sensors  $\downarrow$ & 5.0 & 3.19 & 2.65 & 1.0 & 2.0 & 2.99 & 2.56 & 4.05 & 3.48 & 5.0 & \ourscell{2.08} \\
          \theadfont Energy vs AON (\%)  $\downarrow$ & 100.0 & 61.2 & 49.8 & 24.2 & 38.9 & 60.1 & 91.2 & 68.2 & 67.5 & 100 & \ourscell{42.3} \\
          \theadfont Runtime (ms)  $\downarrow$ & 14.9 & 14.7 & 26.8 & 13.6 & 13.7 & 14.4 & 84.1 & 14.1 & 12.1 & 565.3 & \ourscell{14.3} \\
          \theadfont Peak RAM (MB)  $\downarrow$ & 305 & 282 & 403 & 277 & 281 & 287 & 674 & 279 & 299 & 731 & \ourscell{284} \\
          \bottomrule
        \end{tabularx}
      \endgroup
    \end{minipage}%
  }

\sbox{\RightTbl}{%
  \begin{minipage}[t]{0.28\textwidth}
    \centering
    \begingroup
      \scriptsize
      \setlength{\tabcolsep}{2pt}
      \newcommand{\RightTblRowStretch}{1.13} 
      \renewcommand{\arraystretch}{\RightTblRowStretch}
      \newcolumntype{C}{>{\centering\arraybackslash}X}                           
      \newcolumntype{S}{>{\hsize=.9\hsize\centering\arraybackslash}X}           
      \begin{tabularx}{\linewidth}{@{}l!{\vrule width 0.1pt} S C S @{}}
        \toprule
        {\theadfont\thc{$r$ (m)}} &
        {\theadfont\thc{\baseline{IGG}}} &
        {\theadfont\thc{\baseline{DL}}} &
        {\theadfont\thc{\baseline{\BCOD}}} \\
        \midrule
        25  & 7.5 ms   & 565 ms  & \ourscell{9.8 ms}  \\
        40  & 10.9 ms  & 1446 ms & \ourscell{9.7 ms}  \\
        55  & 14.6 ms  & 2737 ms & \ourscell{9.6 ms}  \\
        70  & 18.2 ms  & 4430 ms & \ourscell{10.7 ms} \\
        85  & 18.7 ms  & 6536 ms & \ourscell{10.4 ms} \\
        100 & 23.3 ms  & 9040 ms & \ourscell{10.9 ms} \\
        \bottomrule
      \end{tabularx}
    \endgroup
  \end{minipage}%
}

  \setlength{\LeftTblHeight}{\dimexpr\ht\LeftTbl+\dp\LeftTbl\relax}

  \begin{subtable}[t]{0.69\textwidth}
    \centering
    \usebox{\LeftTbl}
    \vspace{-0.5pt}
    \caption{\small
      Performance comparison.  
      Keys – AON: \baseline{Always-ON}; GOF: \baseline {Greedy-OFF}; IGG: \baseline{InfoGain-Greedy};
      R1/R2: \baseline{Random 1/2}; $\sigma$M: \baseline{$\sigma$-Mean}; SS: \baseline{Sample-Spread}; NB: \baseline{No-Belief Raster};
      PRL: \baseline{Pure-RL}; DL: \baseline{DESPOT-Lite}}
    \label{tabres}
  \end{subtable}
  \hfill
  \begin{subtable}[t]{0.28\textwidth}
    \centering
    \begin{minipage}[t][\LeftTblHeight][t]{\linewidth}
      \centering
      \usebox{\RightTbl}
    \end{minipage}
    \caption{\small Wall-clock latency vs.\ radius $r$. Average of 5 runs.}
    \label{tab:radius_scaling}
  \end{subtable}

\end{table*}
\endgroup

\kf{Key Finding \#2: B-COD's variance is a calibrated, context-aware predictor of localisation error.}
\vspace{-6pt}
\begin{figure}[H]
    \centering
    \includegraphics[width=0.85\linewidth,trim=30 100 28 35,clip]{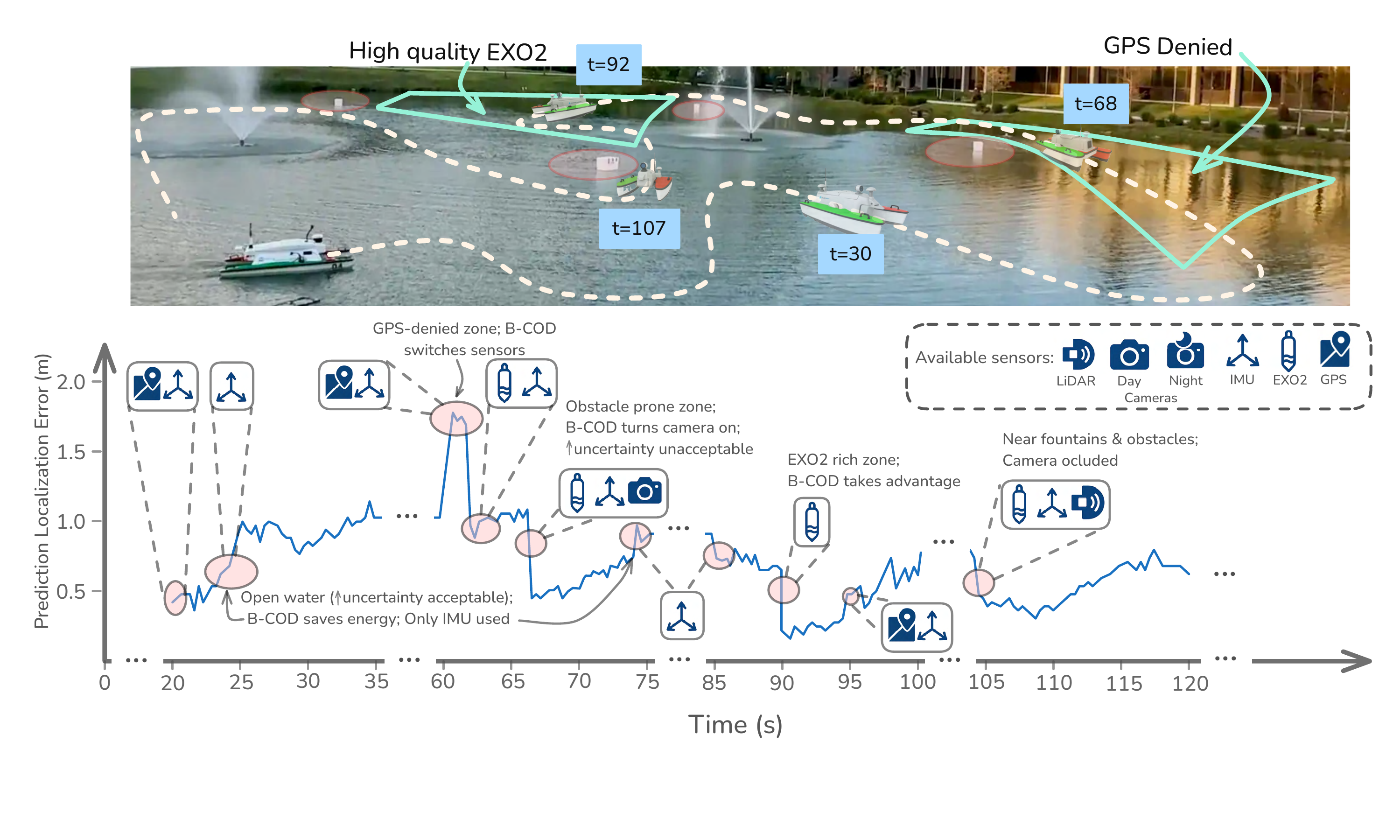}
    \caption{Representative lap with B-COD. Obstacles (red circles), GPS denied and rich EXO2 zones are marked. The lower panel plots B-COD's predicted localisation error vs time alongside the sensor-activation traces (only representative toggles are shown). Brief annotations indicate our interpretation that triggered the toggle.}
    \label{fig:traj}
\end{figure}
\vspace{-6pt}
The reliability curve (Figure \ref{fig:calib}) shows numerical calibration (6 \% mean error). The reason behind the bound relaxing follows directly from what the diffusion planner is told to care about: 
\begin{wrapfigure}[15]{r}{0.30\textwidth} 
  \vspace{-6pt}           
  \centering
\includegraphics[width=\linewidth, trim=60 50 230 25, clip]{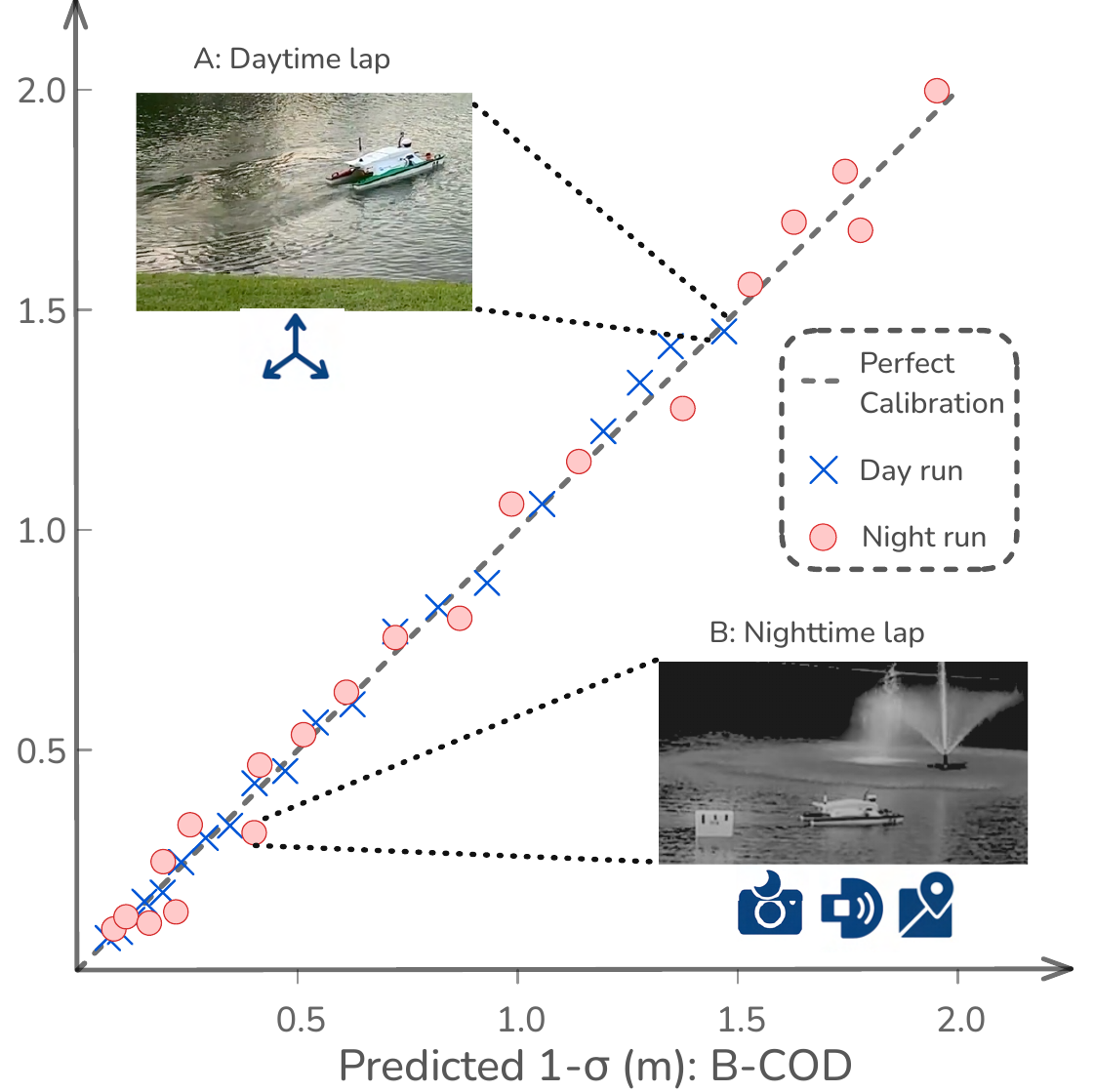}
\caption{\small Reliability diagram: 20 bins plot the empirical planar error (y-axis) against B-COD's predicted (x-axis).}
  \label{fig:calib}
  \vspace{-1100pt}
\end{wrapfigure}
belief shape, active sensors, and local map geometry. 
Call-out B (night lap, obstacle corridor): The semantic map reports obstacles--a fountain and a floating buoy. 
Colliding here would be mission-ending, so B-COD turns on LiDAR, night camera and IMU (high energy).  
The planner now expects rich pose updates and knows that centimeters of pose error matter; it shrinks the bound to 0.45 m, almost matching the 0.46 m ground truth drift.
Call-out A (day lap): Tens of meters separate the ASV from any hazard.  With only the IMU running (low energy), B-COD predicts pure dead-reckoning growth yet also ``knows" that a meter of drift will not intersect anything.  It therefore widens the bound to 1.85 m, closely tracking the ground truth 2.0 m error.
These results demonstrate that $u^{\mathrm{CVaR}}$ is numerically reliable and spatially discriminative, providing the scheduler with the rich information needed to trade energy for certainty.
Figure \ref{fig:traj} shows these risk swings along the full lap for qualitative context.

\kf{Key Finding \#3: B-COD stays within a 10 $\pm$ 1 ms envelope and out-scales analytic belief planners.}
Table~\ref{tab:radius_scaling} sweeps the workspace radius from 25 m to 100 m (full lake sector).  B-COD's latency is flat--10.3 $\pm$ 0.6 ms throughout—because the belief crop is \textit{always} down-sampled and the UNet's receptive field is fixed; compute therefore scales with network width, not with world area.  The InfoGain-Greedy baseline must update an \(n\)-cell covariance grid; its cost grows \(\Theta(R^{2})\)\cite{charrow2015information}, reaching 23 ms at 100 m. DESPOT-Lite's branching factor of the belief tree increases with visible free space; runtime balloons to 9000 ms over the same sweep, far beyond what an embedded loop can absorb. The takeaway is practical as well as theoretical: constant-time scaling lets B-COD replan over lake-scale horizons without ever violating the real-time threshold, whereas analytic planners become the computational bottleneck well before the map reaches lake-scale.

\kf{Key Finding \#4: B-COD adapts online, re-allocating modalities to recover from faults.}
\begin{wrapfigure}[11]{r}{0.33\textwidth} 
  \vspace{-6pt}           
  \centering  
  \includegraphics[width=\linewidth,trim=63 58 60 45,clip]{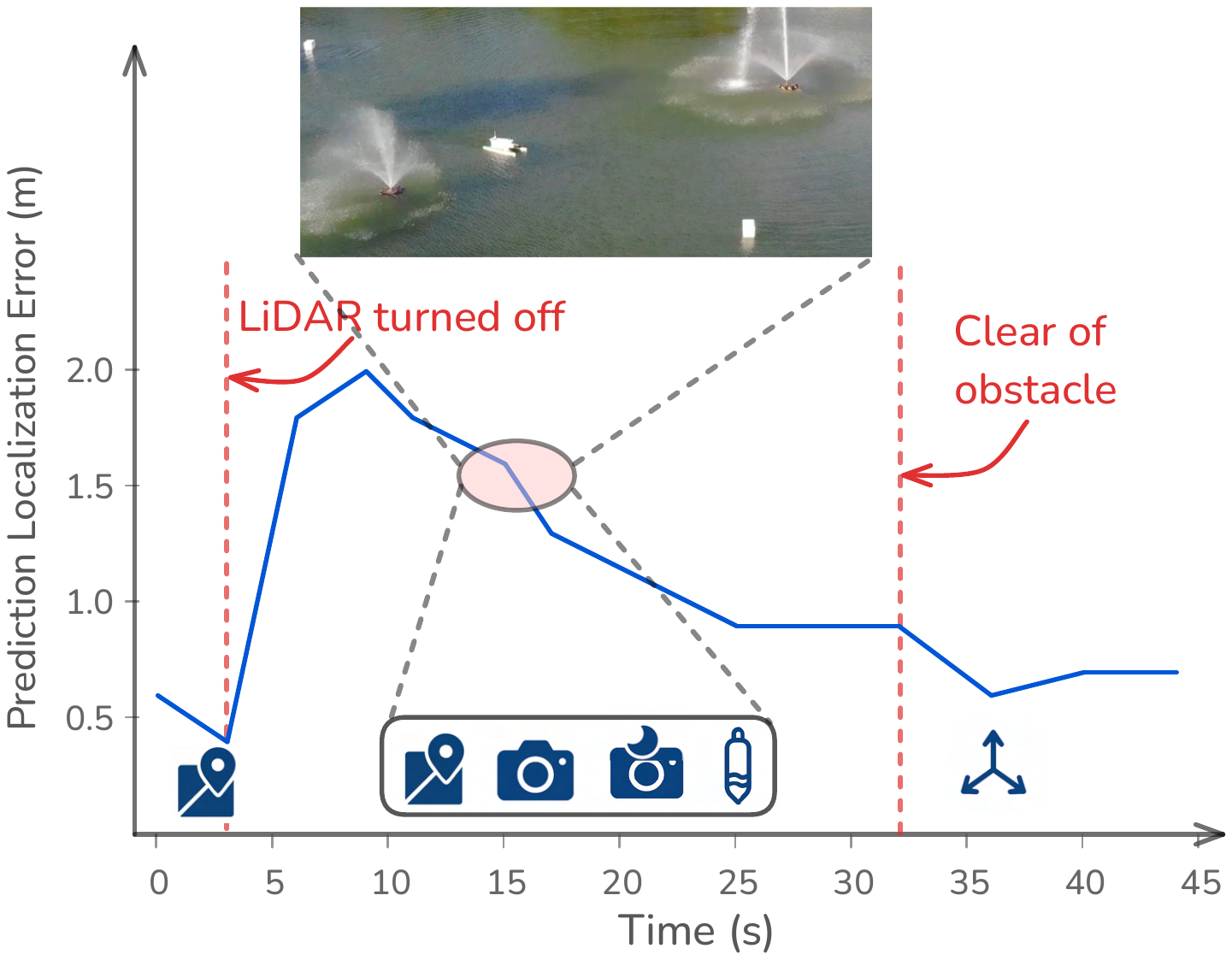}
  \caption{\small Scheduler response to a forced LiDAR outage during a lap.}
  \label{fig:calib_b}
  \vspace{-6pt}            
\end{wrapfigure}
During a daytime lap, we manually disabled the LiDAR 30 s before the ASV entered the narrow fountain corridor, which demands typically sub-meter localisation. B-COD's risk proxy spiked from 0.6 m to 1.8 m as soon as the loss of range data was reflected. The SA reacted on the next cycle: it re-enabled both cameras and the EXO2 sonde, accepted the high energy penalty, and drove risk down to 0.8 m. Once clear of the obstacle, the proxy dropped naturally; the scheduler shut the extra modalities off and returned to the energy-saving IMU-only sensor choice. No heuristic was required--the planner's calibrated variance alone drove the correct, context-specific recovery sequence.

\kf{Ablations: } Removing the belief raster (NB) deprives the planner of map semantics: goal-reach collapses to 68 \%, collisions triple, and the SAC reacts by keeping four sensors on. Replacing CVaR with a naive waypoint average ($\sigma$-M) leaves the variance head intact but blurs risk peaks; the policy over-compensates, averaging three sensors and burning 60\% of Always-ON energy, while still breaching the risk budget 9\% of the time. Estimating risk from 20 Monte-Carlo samples (SS) restores calibration but at the cost of high runtime. 
\vspace{-6pt}
\section{Conclusion}\label{sec:conclusion}
B-COD demonstrates that a diffusion planner can double as a sensing oracle, enabling ``just-enough" sensing for navigation. B-COD couples (i) a belief raster, (ii) a one-step diffusion model conditioned on belief and sensor flags that returns a short trajectory and a calibrated proxy, and (iii) a risk-constrained scheduler that powers only the sensors needed to keep that risk below a user budget. On hardware experiments B-COD reaches its goals with 98\% success while cutting sensing energy by $>$ 50 \%, maintains variance calibration to 6 \%, replans in 10 ms--far faster than analytic belief planners. 

\newpage
\section{Limitations}
Although B-COD delivered strong real-world results--near-perfect goal completion, tight error calibration, and substantial energy savings--it still rests on some assumptions that limit immediate deployment beyond the scenarios tested.

\textit{Domain-specific retraining:}
The sensor mask enters the network as a learned embedding whose dimension and semantics are fixed at training time.  Adding a new modality—or swapping a lidar for a lower-power radar—requires collecting demonstrations and fine-tuning the entire model.  A more versatile alternative is to encode each sensor through a shared description (e.g., field-of-view, expected information gain per joule) and learn the embedding once, allowing plug-and-play expansion of the suite.

\textit{Dependence on a static semantic map:}  
B-COD conditions on pre-loaded layers such as obstacle occupancy and GNSS visibility.  If construction alters the shoreline or a temporary stage blocks a channel, the planner may route through stale free space.  
Coupling the raster with on-line semantic SLAM or incremental map-updating would mitigate this brittleness and let the system react to novel structures and transient occlusions.

\textit{Binary sensor model:}  
The current scheduler flips each modality fully on or off.  In reality many sensors admit graded settings—reduced lidar scan rate, lower camera resolution, duty-cycled GPS fixes—that offer finer energy/performance trade-offs.  Extending the action space to multi-level or continuous controls, possibly guided by differentiable power/quality curves, could unlock further savings without compromising risk.

\textit{Calibration tied to training distribution:  }
Variance honesty rests on the assumption that the demonstration data span the operating envelope.  If the robot later encounters lighting, weather, or sensor faults absent from the corpus, \(u^{\mathrm{CVaR}}\) may under-state true error.  Techniques such as confidence-aware data augmentation, out-of-distribution detectors, or post-deployment Bayesian recalibration would help maintain reliability as conditions drift.

Addressing these limitations—expressive sensor descriptors, live map updates, multi-level energy controls, and distribution-aware calibration—forms the next step toward a fully adaptive, task-agnostic “just-enough sensing’’ navigation stack.

\section{Acknowledgments}
This work was supported by National Science Foundation grants 2118329, IIS-2024733, IIS-2331908, the Air Force Research Laboratory under Award FA8651-23-1-0003,  the Office of Naval Research under the grants N00014-23-1-2789, N00014-25-1-2369, N00014-23-1-2505, and N00014-23-1-2651,  the U.S. Department of Defense grant 78170-RT-REP,  the FDEP grant INV31, and by the Army Research Laboratories under contract W911NF1920243. This is publication \#2035 from the Institute of Environment at Florida International University

\clearpage


\bibliography{example}  

\begin{thebibliography}{117}
\providecommand{\natexlab}[1]{#1}
\providecommand{\url}[1]{\texttt{#1}}
\expandafter\ifx\csname urlstyle\endcsname\relax
  \providecommand{\doi}[1]{doi: #1}\else
  \providecommand{\doi}{doi: \begingroup \urlstyle{rm}\Url}\fi

\bibitem[Cadena et~al.(2016)Cadena, Carlone, Carrillo, Latif, Scaramuzza, Neira, Reid, and Leonard]{cadena2016past}
C.~Cadena, L.~Carlone, H.~Carrillo, Y.~Latif, D.~Scaramuzza, J.~Neira, I.~Reid, and J.~J. Leonard.
\newblock Past, present, and future of simultaneous localization and mapping{: T}oward the robust-perception age.
\newblock \emph{{IEEE Transactions on Robotics}}, 2016.

\bibitem[Debeunne and Vivet(2020)]{debeunne2020review}
C.~Debeunne and D.~Vivet.
\newblock A review of visual-{LiDAR} fusion based simultaneous localization and mapping.
\newblock \emph{Sensors}, 2020.

\bibitem[Wang et~al.(2020)Wang, Wu, and Niu]{wang2020multisensor}
Z.~Wang, Y.~Wu, and Q.~Niu.
\newblock Multi-sensor fusion in automated driving{: A} survey.
\newblock \emph{{IEEE Access}}, 2020.

\bibitem[Majumdar et~al.(2023)Majumdar, Mei, and Pacelli]{majumdar2023fundamental}
A.~Majumdar, Z.~Mei, and V.~Pacelli.
\newblock Fundamental limits for sensor-based robot control.
\newblock \emph{The International Journal of Robotics Research}, 2023.

\bibitem[Malawade et~al.(2022)Malawade, Mortlock, and Al~Faruque]{malawade2022ecofusion}
A.~V. Malawade, T.~Mortlock, and M.~A. Al~Faruque.
\newblock {EcoFusion: E}nergy-aware adaptive sensor fusion for efficient autonomous vehicle perception.
\newblock In \emph{ACM/IEEE Design Automation Conference}, 2022.

\bibitem[Kim et~al.(2018)Kim, Park, and Chung]{kim2018selfdiagnosis}
J.~Kim, J.~Park, and W.~Chung.
\newblock Self-diagnosis of localization status for autonomous mobile robots.
\newblock \emph{Sensors}, 2018.

\bibitem[Aqel et~al.(2016)Aqel, Marhaban, Saripan, and Ismail]{aqel2016vo_review}
M.~O.~A. Aqel, M.~H. Marhaban, M.~I. Saripan, and N.~B. Ismail.
\newblock Review of visual odometry: Types, approaches, challenges, and applications.
\newblock \emph{SpringerPlus}, 2016.

\bibitem[akbar Agha-mohammadi et~al.(2014)akbar Agha-mohammadi, Chakravorty, and Amato]{rw2}
A.~akbar Agha-mohammadi, S.~Chakravorty, and N.~M. Amato.
\newblock {FIRM: S}ampling-based feedback motion-planning under motion uncertainty and imperfect measurements.
\newblock \emph{The International Journal of Robotics Research}, 2014.

\bibitem[van~den Berg et~al.(2010)van~den Berg, Abbeel, and Goldberg]{berg2010lqgmp}
J.~van~den Berg, P.~Abbeel, and K.~Goldberg.
\newblock {LQG-MP: O}ptimized path planning for robots with motion uncertainty and imperfect state information.
\newblock In \emph{Robotics: Science and Systems}, 2010.

\bibitem[Barbosa et~al.(2021)Barbosa, Lacerda, Duckworth, Tumova, and Hawes]{BarbosaRiskAware}
F.~S. Barbosa, B.~Lacerda, P.~Duckworth, J.~Tumova, and N.~Hawes.
\newblock Risk-aware motion planning in partially known environments.
\newblock In \emph{IEEE Conference on Decision and Control}, 2021.

\bibitem[Papachristos et~al.(2017)Papachristos, Khattak, and Alexis]{papachristos2017autonomous}
C.~Papachristos, S.~Khattak, and K.~Alexis.
\newblock Autonomous exploration of visually-degraded environments using aerial robots.
\newblock In \emph{IEEE International Conference on Unmanned Aircraft Systems}, 2017.

\bibitem[Ondruška et~al.(2015)Ondruška, Gurău, Marchegiani, Tong, and Posner]{ondruska2015scheduled}
P.~Ondruška, C.~Gurău, L.~Marchegiani, C.~H. Tong, and I.~Posner.
\newblock Scheduled perception for energy-efficient path following.
\newblock In \emph{IEEE International Conference on Robotics and Automation}, 2015.

\bibitem[Kaelbling et~al.(1998)Kaelbling, Littman, and Cassandra]{kaelbling1998planning}
L.~P. Kaelbling, M.~L. Littman, and A.~R. Cassandra.
\newblock Planning and acting in partially observable stochastic domains.
\newblock \emph{Artificial Intelligence}, 1998.

\bibitem[Spaan and Lima(2009)]{spaan2009dynamic}
M.~T.~J. Spaan and P.~U. Lima.
\newblock A decision-theoretic approach to dynamic sensor selection in camera networks.
\newblock In \emph{International Conference on Automated Planning and Scheduling}, 2009.

\bibitem[Charrow et~al.(2015)Charrow, Michael, and Kumar]{charrow2015information}
B.~Charrow, N.~Michael, and V.~Kumar.
\newblock Information-theoretic planning with trajectory optimization for dense 3d mapping.
\newblock In \emph{Robotics: Science and Systems}, 2015.

\bibitem[Leong et~al.(2020)Leong, Ramaswamy, Quevedo, Karl, and Shi]{leong2020deep}
A.~S. Leong, A.~Ramaswamy, D.~E. Quevedo, H.~Karl, and L.~Shi.
\newblock Deep reinforcement learning for wireless sensor scheduling in cyber-physical systems.
\newblock \emph{Automatica}, 2020.

\bibitem[Alali et~al.(2024)Alali, Kazeminajafabadi, and Imani]{alali2024drlschedule}
M.~Alali, A.~Kazeminajafabadi, and M.~Imani.
\newblock Deep reinforcement learning sensor scheduling for effective monitoring of dynamical systems.
\newblock \emph{Systems Science \& Control Engineering}, 2024.

\bibitem[Puthumanaillam et~al.(2024)Puthumanaillam, Vora, and Ornik]{comtraqmpc}
G.~Puthumanaillam, M.~Vora, and M.~Ornik.
\newblock Comtraq-mpc: Meta-trained dqn-mpc integration for trajectory tracking with limited active localization updates.
\newblock In \emph{2024 IEEE/RSJ International Conference on Intelligent Robots and Systems (IROS)}, pages 13592--13598, 2024.
\newblock \doi{10.1109/IROS58592.2024.10801659}.

\bibitem[Akkaya et~al.(2019)Akkaya, Andrychowicz, Chociej, Litwin, McGrew, Petron, Paino, Plappert, Powell, Ribas, et~al.]{akkaya2019solving}
I.~Akkaya, M.~Andrychowicz, M.~Chociej, M.~Litwin, B.~McGrew, A.~Petron, A.~Paino, M.~Plappert, G.~Powell, R.~Ribas, et~al.
\newblock Solving rubik's cube with a robot hand.
\newblock \emph{arXiv preprint arXiv:1910.07113}, 2019.

\bibitem[Alshiekh et~al.(2018)Alshiekh, Bloem, Ehlers, K{\"o}nighofer, Niekum, Topcu, and Wulf]{alshiekh2018safe}
M.~Alshiekh, R.~Bloem, R.~Ehlers, B.~K{\"o}nighofer, S.~Niekum, U.~Topcu, and M.~Wulf.
\newblock Safe reinforcement learning via shielding.
\newblock In \emph{AAAI Conference on Artificial Intelligence}, 2018.

\bibitem[Honda et~al.(2024)Honda, Yonetani, Nishimura, and Kozuno]{honda2024whentoreplan}
K.~Honda, R.~Yonetani, M.~Nishimura, and T.~Kozuno.
\newblock When to replan? {An} adaptive replanning strategy for autonomous navigation using deep reinforcement learning.
\newblock In \emph{IEEE International Conference on Robotics and Automation}, 2024.

\bibitem[Janner et~al.(2022)Janner, Du, Tenenbaum, and Levine]{janner2022diffuser}
M.~Janner, Y.~Du, J.~Tenenbaum, and S.~Levine.
\newblock Planning with diffusion for flexible behavior synthesis.
\newblock In \emph{International Conference on Machine Learning}, 2022.

\bibitem[Luo et~al.(2024)Luo, Sun, Tenenbaum, and Du]{luo2024potential}
Y.~Luo, C.~Sun, J.~B. Tenenbaum, and Y.~Du.
\newblock Potential based diffusion motion planning.
\newblock In \emph{International Conference on Machine Learning}, 2024.

\bibitem[Chi et~al.(2023)Chi, Feng, Du, Xu, Cousineau, Burchfiel, and Song]{chi2023diffusionpolicy}
C.~Chi, S.~Feng, Y.~Du, Z.~Xu, E.~Cousineau, B.~Burchfiel, and S.~Song.
\newblock Diffusion policy{: V}isuomotor policy learning via action diffusion.
\newblock In \emph{Robotics: Science and Systems}, 2023.

\bibitem[Carvalho et~al.(2023)Carvalho, Le, Baierl, Koert, and Peters]{carvalho2023mpdiffusion}
J.~a. Carvalho, A.~T. Le, M.~Baierl, D.~Koert, and J.~Peters.
\newblock Motion planning diffusion{: L}earning and planning of robot motions with diffusion models.
\newblock In \emph{IEEE/RSJ International Conference on Intelligent Robots and Systems}, 2023.

\bibitem[Sun et~al.(2023)Sun, Jiang, Qiu, Talpur~Nobel, Kochenderfer, and Schwager]{sun2023plancp}
J.~Sun, Y.~Jiang, J.~Qiu, P.~Talpur~Nobel, M.~Kochenderfer, and M.~Schwager.
\newblock Conformal prediction for uncertainty-aware planning with diffusion dynamics models.
\newblock In \emph{Advances in Neural Information Processing Systems}, 2023.

\bibitem[Ho et~al.(2020)Ho, Jain, and Abbeel]{ho2020denoising}
J.~Ho, A.~Jain, and P.~Abbeel.
\newblock Denoising diffusion probabilistic models.
\newblock In \emph{Advances in Neural Information Processing Systems}, 2020.

\bibitem[Feng et~al.(2024)Feng, Gu, An, and Pan]{feng2024tat}
L.~Feng, P.~Gu, B.~An, and G.~Pan.
\newblock Resisting stochastic risks in diffusion planners with the trajectory aggregation tree.
\newblock In \emph{International Conference on Machine Learning}, 2024.

\bibitem[Berg et~al.(2010)Berg, Abbeel, and Goldberg]{rw1}
J.~V.~D. Berg, P.~Abbeel, and K.~Goldberg.
\newblock Lqg-mp: Optimized path planning for robots with motion uncertainty and imperfect state information.
\newblock In \emph{Robotics: Science and Systems}, 2010.

\bibitem[Sun and Kumar(2021)]{rw3}
K.~Sun and V.~Kumar.
\newblock Belief space planning for mobile robots with range sensors using ilqg.
\newblock \emph{IEEE Robotics and Automation Letters}, 2021.

\bibitem[Thrun et~al.(2004)Thrun, Liu, Koller, Ng, Ghahramani, and Durrant-Whyte]{a1}
S.~Thrun, Y.~Liu, D.~Koller, A.~Y. Ng, Z.~Ghahramani, and H.~Durrant-Whyte.
\newblock Simultaneous localization and mapping with sparse extended information filters.
\newblock \emph{The International Journal of Robotics Research}, 2004.

\bibitem[Shan et~al.(2020)Shan, Englot, Meyers, Wang, Ratti, and Rus]{a2}
T.~Shan, B.~Englot, D.~Meyers, W.~Wang, C.~Ratti, and D.~Rus.
\newblock {LIO-SAM: T}ightly-coupled lidar inertial odometry via smoothing and mapping.
\newblock In \emph{IEEE/RSJ International Conference on Intelligent Robots and Systems}, 2020.

\bibitem[Dellaert and Kaess(2006)]{a3}
F.~Dellaert and M.~Kaess.
\newblock {Square root SAM: S}imultaneous localization and mapping via square root information smoothing.
\newblock \emph{The International Journal of Robotics Research}, pages 1181--1203, 2006.

\bibitem[Kaess et~al.(2008)Kaess, Ranganathan, and Dellaert]{a4}
M.~Kaess, A.~Ranganathan, and F.~Dellaert.
\newblock {iSAM: I}ncremental smoothing and mapping.
\newblock \emph{IEEE Transactions on Robotics}, 2008.

\bibitem[Rosinol et~al.(2021)Rosinol, Violette, Abate, Hughes, Chang, Shi, Gupta, and Carlone]{a5}
A.~Rosinol, A.~Violette, M.~Abate, N.~Hughes, Y.~Chang, J.~Shi, A.~Gupta, and L.~Carlone.
\newblock {Kimera: From SLAM} to spatial perception with {3D} dynamic scene graphs.
\newblock \emph{The International Journal of Robotics Research}, 2021.

\bibitem[Karkus et~al.(2021)Karkus, Cai, and Hsu]{a6}
P.~Karkus, S.~Cai, and D.~Hsu.
\newblock Differentiable {SLAM-Net: L}earning particle slam for visual navigation.
\newblock In \emph{IEEE/CVF Conference on Computer Vision and Pattern Recognition}, 2021.

\bibitem[Younis and Sudderth(2024)]{a7}
A.~Younis and E.~Sudderth.
\newblock {Learning to be Smooth: An }end-to-end differentiable particle smoother.
\newblock In \emph{Neural Information Processing Systems}, 2024.

\bibitem[Lim and Chon(2024)]{a8}
J.~Lim and K.~H. Chon.
\newblock {Minimax Rao-blackwellized} particle filtering in {2D LIDAR SLAM}.
\newblock \emph{International Journal of Control, Automation and Systems}, 2024.

\bibitem[Liu et~al.(2024)Liu, Xu, Qiao, Jiang, and Yu]{a9}
T.~Liu, C.~Xu, Y.~Qiao, C.~Jiang, and J.~Yu.
\newblock Particle filter {SLAM} for vehicle localization.
\newblock \emph{arXiv preprint arXiv:2402.07429}, 2024.

\bibitem[Pineau et~al.(2003)Pineau, Gordon, Thrun, et~al.]{b1}
J.~Pineau, G.~Gordon, S.~Thrun, et~al.
\newblock Point-based value {iteration: An anytime algorithm for POMDPs}.
\newblock In \emph{International Joint Conference on Artificial Intelligence}, 2003.

\bibitem[Spaan and Vlassis(2005)]{b2}
M.~T. Spaan and N.~Vlassis.
\newblock {Perseus: Randomized} point-based value iteration for {POMDPs}.
\newblock \emph{Journal of Artificial Intelligence Research}, 2005.

\bibitem[Kurniawati et~al.(2008)Kurniawati, Hsu, and Lee]{b3}
H.~Kurniawati, D.~Hsu, and W.~S. Lee.
\newblock {SARSOP: Efficient} point-based pomdp planning by approximating optimally reachable belief spaces.
\newblock In \emph{Robotics: Science and systems}, volume 2008. Citeseer, 2008.

\bibitem[Somani et~al.(2013)Somani, Ye, Hsu, and Lee]{b4}
A.~Somani, N.~Ye, D.~Hsu, and W.~S. Lee.
\newblock {DESPOT: Online POMDP} planning with regularization.
\newblock In \emph{Advances in Neural Information Processing Systems}, 2013.

\bibitem[Cai et~al.(2021)Cai, Luo, Hsu, and Lee]{b5}
P.~Cai, Y.~Luo, D.~Hsu, and W.~S. Lee.
\newblock {HyP-DESPOT: A} hybrid parallel algorithm for online planning under uncertainty.
\newblock \emph{The International Journal of Robotics Research}, 2021.

\bibitem[Liang et~al.(2024)Liang, Kim, Thomason, Kingston, Kurniawati, and Kavraki]{b6}
Y.~Liang, E.~Kim, W.~Thomason, Z.~Kingston, H.~Kurniawati, and L.~E. Kavraki.
\newblock Scaling long-horizon online {POMDP} planning via rapid state space sampling.
\newblock \emph{arXiv preprint arXiv:2411.07032}, 2024.

\bibitem[Du~Toit and Burdick(2011)]{b7}
N.~E. Du~Toit and J.~W. Burdick.
\newblock Probabilistic collision checking with chance constraints.
\newblock \emph{IEEE Transactions on Robotics}, 2011.

\bibitem[Luders et~al.(2010)Luders, Kothari, and How]{b8}
B.~Luders, M.~Kothari, and J.~How.
\newblock Chance constrained {RRT} for probabilistic robustness to environmental uncertainty.
\newblock In \emph{AIAA Guidance, Navigation, and Control Conference}, 2010.

\bibitem[Dai et~al.(2019)Dai, Schaffert, Jasour, Hofmann, and Williams]{b9}
S.~Dai, S.~Schaffert, A.~Jasour, A.~Hofmann, and B.~Williams.
\newblock Chance constrained motion planning for high-dimensional robots.
\newblock In \emph{International Conference on Robotics and Automation}, 2019.

\bibitem[Glasheen et~al.(2024)Glasheen, Bird, and Frew]{b10}
K.~Glasheen, J.~J. Bird, and E.~W. Frew.
\newblock Experimental assessment of chance-constrained motion planning for small uncrewed aircraft.
\newblock \emph{Field Robotics}, 2024.

\bibitem[Bourgault et~al.(2002)Bourgault, Makarenko, Williams, Grocholsky, and Durrant-Whyte]{c1}
F.~Bourgault, A.~A. Makarenko, S.~B. Williams, B.~Grocholsky, and H.~F. Durrant-Whyte.
\newblock Information based adaptive robotic exploration.
\newblock In \emph{IEEE/RSJ International Conference on Intelligent Robots and Systems}, 2002.

\bibitem[Miller and Murphey(2013)]{c2}
L.~M. Miller and T.~D. Murphey.
\newblock Trajectory optimization for continuous ergodic exploration.
\newblock In \emph{American Control Conference}, 2013.

\bibitem[Charrow et~al.(2015)Charrow, Kahn, Patil, Liu, Goldberg, Abbeel, Michael, and Kumar]{c3}
B.~Charrow, G.~Kahn, S.~Patil, S.~Liu, K.~Goldberg, P.~Abbeel, N.~Michael, and V.~Kumar.
\newblock Information-theoretic planning with trajectory optimization for dense {3D} mapping.
\newblock In \emph{Robotics: Science and Systems}, 2015.

\bibitem[Sun et~al.(2024)Sun, Gaggar, Trautman, and Murphey]{c4}
M.~M. Sun, A.~Gaggar, P.~Trautman, and T.~Murphey.
\newblock Fast ergodic search with kernel functions.
\newblock \emph{arXiv preprint arXiv:2403.01536}, 2024.

\bibitem[Dutta et~al.(2023)Dutta, Wilde, and Smith]{d1}
S.~Dutta, N.~Wilde, and S.~L. Smith.
\newblock A unified approach to optimally solving sensor scheduling and sensor selection problems in {Kalman} filtering.
\newblock In \emph{IEEE Conference on Decision and Control}, 2023.

\bibitem[Carlone and Lyons(2014)]{d2}
L.~Carlone and D.~Lyons.
\newblock Uncertainty-constrained robot exploration{: A} mixed-integer linear programming approach.
\newblock In \emph{IEEE International Conference on Robotics and Automation}, 2014.

\bibitem[Schlotfeldt et~al.(2021)Schlotfeldt, Tzoumas, and Pappas]{d3}
B.~Schlotfeldt, V.~Tzoumas, and G.~J. Pappas.
\newblock Resilient active information acquisition with teams of robots.
\newblock \emph{IEEE Transactions on Robotics}, 2021.

\bibitem[Yu et~al.(2014)Yu, Schwager, and Rus]{d4}
J.~Yu, M.~Schwager, and D.~Rus.
\newblock Correlated orienteering problem and its application to informative path planning for persistent monitoring tasks.
\newblock In \emph{IEEE/RSJ International Conference on Intelligent Robots and Systems}, 2014.

\bibitem[Lauri et~al.(2020)Lauri, Pajarinen, Peters, and Frintrop]{d5}
M.~Lauri, J.~Pajarinen, J.~Peters, and S.~Frintrop.
\newblock Multi-sensor next-best-view planning as matroid-constrained submodular maximization.
\newblock \emph{IEEE Robotics and Automation Letters}, 2020.

\bibitem[Corah and Michael(2019)]{d6}
M.~Corah and N.~Michael.
\newblock Distributed matroid-constrained submodular maximization for multi-robot exploration: Theory and practice.
\newblock \emph{Autonomous Robots}, 2019.

\bibitem[Kazma and Taha(2024)]{d7}
M.~H. Kazma and A.~F. Taha.
\newblock Multilinear extensions in submodular optimization for optimal sensor scheduling in nonlinear networks.
\newblock \emph{arXiv preprint arXiv:2408.03833}, 2024.

\bibitem[Maity and Baras(2015)]{e1}
D.~Maity and J.~S. Baras.
\newblock Dynamic, optimal sensor scheduling and value of information.
\newblock In \emph{International Conference on Information Fusion}, 2015.

\bibitem[Bui et~al.(2023)Bui, Pandey, Chiariotti, and Popovski]{e2}
V.-P. Bui, S.~R. Pandey, F.~Chiariotti, and P.~Popovski.
\newblock Scheduling policy for value-of-information ({VOI}) in trajectory estimation for digital twins.
\newblock \emph{IEEE Communications Letters}, 2023.

\bibitem[Krishna et~al.(2025)Krishna, Hu, and Jayaraman]{e4}
A.~Krishna, E.~S. Hu, and D.~Jayaraman.
\newblock The value of sensory information to a robot.
\newblock In \emph{International Conference on Learning Representations}, 2025.

\bibitem[Jawaid and Smith(2015)]{e5}
S.~T. Jawaid and S.~L. Smith.
\newblock Submodularity and greedy algorithms in sensor scheduling for linear dynamical systems.
\newblock \emph{Automatica}, 2015.

\bibitem[Hashemi et~al.(2018)Hashemi, Ghasemi, Vikalo, and Topcu]{e6}
A.~Hashemi, M.~Ghasemi, H.~Vikalo, and U.~Topcu.
\newblock A randomized greedy algorithm for near-optimal sensor scheduling in large-scale sensor networks.
\newblock In \emph{American Control Conference}, 2018.

\bibitem[Liu et~al.(2022)Liu, Li, Johansson, M{\aa}rtensson, and Xie]{e7}
H.~Liu, Y.~Li, K.~H. Johansson, J.~M{\aa}rtensson, and L.~Xie.
\newblock Rollout approach to sensor scheduling for remote state estimation under integrity attack.
\newblock \emph{Automatica}, 2022.

\bibitem[Hoffmann et~al.(2021)Hoffmann, Charlish, Ritchie, and Griffiths]{e8}
F.~Hoffmann, A.~Charlish, M.~Ritchie, and H.~Griffiths.
\newblock Policy rollout action selection in continuous domains for sensor path planning.
\newblock \emph{IEEE Transactions on Aerospace and Electronic Systems}, 2021.

\bibitem[Jasim(2020)]{e9}
H.~A. Jasim.
\newblock \emph{Dynamic duty cycle mechnism for mobility in wirless sensor networks}.
\newblock PhD thesis, Universiti Tun Hussein Onn Malaysia, 2020.

\bibitem[Diyan et~al.(2020)Diyan, Khan, Nathali~Silva, and Han]{e10}
M.~Diyan, M.~Khan, B.~Nathali~Silva, and K.~Han.
\newblock Scheduling sensor duty cycling based on event detection using bi-directional long short-term memory and reinforcement learning.
\newblock \emph{Sensors}, 2020.

\bibitem[Alali et~al.(2024)Alali, Kazeminajafabadi, and Imani]{f1}
M.~Alali, A.~Kazeminajafabadi, and M.~Imani.
\newblock Deep reinforcement learning sensor scheduling for effective monitoring of dynamical systems.
\newblock \emph{Systems Science \& Control Engineering}, 2024.

\bibitem[Chen et~al.(2023)Chen, Liu, Quevedo, Khosravirad, Li, and Vucetic]{f2}
J.~Chen, W.~Liu, D.~E. Quevedo, S.~R. Khosravirad, Y.~Li, and B.~Vucetic.
\newblock Structure-enhanced drl for optimal transmission scheduling.
\newblock \emph{IEEE Transactions on Wireless Communications}, 2023.

\bibitem[Hajiakhondi-Meybodi et~al.(2022)Hajiakhondi-Meybodi, Hou, and Mohammadi]{f3}
Z.~Hajiakhondi-Meybodi, M.~Hou, and A.~Mohammadi.
\newblock {JUNO: J}ump-start reinforcement learning-based node selection for uwb indoor localization.
\newblock In \emph{IEEE Global Communications Conference}, 2022.

\bibitem[Shurrab et~al.(2022)Shurrab, Singh, Mizouni, and Otrok]{f4}
M.~Shurrab, S.~Singh, R.~Mizouni, and H.~Otrok.
\newblock {IoT} sensor selection for target localization{: A} reinforcement learning based approach.
\newblock \emph{Ad Hoc Networks}, 2022.

\bibitem[Jiang et~al.(2024)Jiang, Chen, Wang, and Xiao]{f5}
C.~Jiang, W.~Chen, Z.~Wang, and W.~Xiao.
\newblock Deep reinforcement learning-based joint sequence scheduling and trajectory planning in wireless rechargeable sensor networks.
\newblock \emph{IEEE Sensors Journal}, 2024.

\bibitem[Bukhari and Kim(2023)]{f6}
A.~Bukhari and H.~Kim.
\newblock Learning-based sensor scheduling for event classification on embedded edge devices.
\newblock In \emph{ACM/IEEE Conference on Internet of Things Design and Implementation}, 2023.

\bibitem[Jurdak et~al.(2010)Jurdak, Corke, Dharman, and Salagnac]{f9}
R.~Jurdak, P.~Corke, D.~Dharman, and G.~Salagnac.
\newblock Adaptive {GPS} duty cycling and radio ranging for energy-efficient localization.
\newblock In \emph{Proceedings of the 8th ACM Conference on Embedded Networked Sensor Systems}, 2010.

\bibitem[Lin et~al.(2010)Lin, Kansal, Lymberopoulos, and Zhao]{f8}
K.~Lin, A.~Kansal, D.~Lymberopoulos, and F.~Zhao.
\newblock Energy-accuracy aware localization for mobile devices.
\newblock 2010.

\bibitem[Ross et~al.(2010)Ross, Gordon, and Bagnell]{g1}
S.~Ross, G.~J. Gordon, and J.~A. Bagnell.
\newblock A reduction of imitation learning and structured prediction to no-regret online learning.
\newblock In \emph{International Conference on Artificial Intelligence and Statistics}, 2010.

\bibitem[Ho and Ermon(2016)]{g2}
J.~Ho and S.~Ermon.
\newblock Generative adversarial imitation learning.
\newblock In \emph{Neural Information Processing Systems}, 2016.

\bibitem[Codevilla et~al.(2017)Codevilla, M{\"u}ller, Dosovitskiy, L{\'o}pez, and Koltun]{g3}
F.~Codevilla, M.~M{\"u}ller, A.~Dosovitskiy, A.~M. L{\'o}pez, and V.~Koltun.
\newblock End-to-end driving via conditional imitation learning.
\newblock \emph{2018 IEEE International Conference on Robotics and Automation (ICRA)}, 2017.

\bibitem[Hepburn and Montana(2022)]{g4}
C.~A. Hepburn and G.~Montana.
\newblock Model-based trajectory stitching for improved behavioural cloning and its applications.
\newblock \emph{ArXiv}, 2022.

\bibitem[Memmel et~al.(2024)Memmel, Berg, Chen, Gupta, and Francis]{g5}
M.~Memmel, J.~Berg, B.~Chen, A.~Gupta, and J.~Francis.
\newblock {STRAP: R}obot sub-trajectory retrieval for augmented policy learning.
\newblock \emph{arXiv preprint arXiv:2412.15182}, 2024.

\bibitem[Lee et~al.(2017)Lee, Choi, Vernaza, Choy, Torr, and Chandraker]{g6}
N.~Lee, W.~Choi, P.~Vernaza, C.~B. Choy, P.~H.~S. Torr, and M.~Chandraker.
\newblock Desire: Distant future prediction in dynamic scenes with interacting agents.
\newblock In \emph{IEEE Conference on Computer Vision and Pattern Recognition}, 2017.

\bibitem[Ivanovic and Pavone(2018)]{g7}
B.~Ivanovic and M.~Pavone.
\newblock The {Trajectron: P}robabilistic multi-agent trajectory modeling with dynamic spatiotemporal graphs.
\newblock In \emph{IEEE/CVF International Conference on Computer Vision}, 2018.

\bibitem[Salzmann et~al.(2020)Salzmann, Ivanovic, Chakravarty, and Pavone]{g8}
T.~Salzmann, B.~Ivanovic, P.~Chakravarty, and M.~Pavone.
\newblock Trajectron++: Dynamically-feasible trajectory forecasting with heterogeneous data.
\newblock In \emph{European Conference on Computer Vision}, 2020.

\bibitem[Janjo{\v{s}} et~al.(2024)Janjo{\v{s}}, Hallgarten, Knittel, Dolgov, Zell, and Z{\"o}llner]{g9}
F.~Janjo{\v{s}}, M.~Hallgarten, A.~Knittel, M.~Dolgov, A.~Zell, and J.~M. Z{\"o}llner.
\newblock Conditional unscented autoencoders for trajectory prediction.
\newblock In \emph{European Conference on Computer Vision}, 2024.

\bibitem[Dong et~al.(2024)Dong, Hao, Yuan, Ni, Wang, Li, and Zheng]{h1}
Z.~Dong, J.~Hao, Y.~Yuan, F.~Ni, Y.~Wang, P.~Li, and Y.~Zheng.
\newblock {DiffuserLite: T}owards real-time diffusion planning.
\newblock In \emph{Advances in Neural Information Processing Systems}, 2024.

\bibitem[Shaoul et~al.(2024)Shaoul, Mishani, Vats, Li, and Likhachev]{h2}
Y.~Shaoul, I.~Mishani, S.~Vats, J.~Li, and M.~Likhachev.
\newblock Multi-robot motion planning with diffusion models.
\newblock \emph{arXiv preprint arXiv:2410.03072}, 2024.

\bibitem[Zhu et~al.(2023)Zhu, Ye, Zhang, Zhao, and Yu]{rw7}
Y.~Zhu, Y.~Ye, S.~Zhang, X.~Zhao, and J.~J. Yu.
\newblock {DiffTraj: G}enerating {GPS} trajectory with diffusion probabilistic model.
\newblock In \emph{Advances in Neural Information Processing Systems}, 2023.

\bibitem[Bouvier et~al.(2025)Bouvier, Ryu, Nagpal, Liao, Sreenath, and Mehr]{bouvier2025ddat}
J.-B. Bouvier, K.~Ryu, K.~Nagpal, Q.~Liao, K.~Sreenath, and N.~Mehr.
\newblock {DDAT: D}iffusion policies enforcing dynamically admissible robot trajectories.
\newblock \emph{arXiv preprint arXiv:2502.15043}, 2025.

\bibitem[Mu et~al.(2015)Mu, Giamou, Paull, akbar Agha-mohammadi, Leonard, and How]{h4}
B.~Mu, M.~Giamou, L.~Paull, A.~akbar Agha-mohammadi, J.~J. Leonard, and J.~P. How.
\newblock Information-based active {SLAM} via topological feature graphs.
\newblock In \emph{IEEE Conference on Decision and Control}, 2015.

\bibitem[Wang et~al.(2022)Wang, Chen, Zhang, and Lou]{h5}
Z.~Wang, H.~Chen, S.~Zhang, and Y.~Lou.
\newblock Active view planning for visual slam in outdoor environments based on continuous information modeling.
\newblock \emph{IEEE/ASME Transactions on Mechatronics}, 2022.

\bibitem[Bryson and Sukkarieh(2008)]{h3}
M.~Bryson and S.~Sukkarieh.
\newblock Observability analysis and active control for airborne {SLAM}.
\newblock \emph{IEEE Transactions on Aerospace and Electronic Systems}, 2008.

\bibitem[Krause et~al.(2005)Krause, Singh, and Guestrin]{h6}
A.~Krause, A.~P. Singh, and C.~Guestrin.
\newblock Near-optimal sensor placements in gaussian processes.
\newblock In \emph{International Conference on Machine Learning}, 2005.

\bibitem[Corah and Michael(2018)]{h7}
M.~Corah and N.~Michael.
\newblock Distributed matroid-constrained submodular maximization for multi-robot exploration{: T}heory and practice.
\newblock \emph{Autonomous Robots}, 2018.

\bibitem[Chaplot et~al.(2020)Chaplot, Gandhi, Gupta, Gupta, and Salakhutdinov]{h8}
D.~S. Chaplot, D.~Gandhi, S.~Gupta, A.~Gupta, and R.~Salakhutdinov.
\newblock Learning to explore using active neural {SLAM}.
\newblock \emph{arXiv preprint arXiv:2004.05155}, 2020.

\bibitem[Yang et~al.(2023)Yang, Liu, Koga, Asgharivaskasi, and Atanasov]{h9}
P.~Yang, Y.~Liu, S.~Koga, A.~Asgharivaskasi, and N.~Atanasov.
\newblock Learning continuous control policies for information-theoretic active perception.
\newblock In \emph{International Conference on Robotics and Automation}, 2023.

\bibitem[Nguyen et~al.(2022)Nguyen, Fyhn, De~Petris, and Alexis]{nguyenmotionprim2022}
H.~Nguyen, S.~H. Fyhn, P.~De~Petris, and K.~Alexis.
\newblock Motion primitives-based navigation planning using deep collision prediction.
\newblock In \emph{International Conference on Robotics and Automation}, 2022.

\bibitem[Dellaert et~al.(1999)Dellaert, Fox, Burgard, and Thrun]{dellaert_monte_1999}
F.~Dellaert, D.~Fox, W.~Burgard, and S.~Thrun.
\newblock Monte {Carlo} localization for mobile robots.
\newblock In \emph{IEEE International Conference on Robotics and Automation}, 1999.

\bibitem[zha(2024)]{zhao_review_2024}
A review of convolutional neural networks in computer vision.
\newblock \emph{Artificial Intelligence Review}, 2024.

\bibitem[Chen et~al.(2024)Chen, Mei, Fan, and Wang]{Chen2024AnOO}
M.~Chen, S.~Mei, J.~Fan, and M.~Wang.
\newblock An overview of diffusion models{: A}pplications, guided generation, statistical rates and optimization.
\newblock \emph{arXiv preprint arXiv:2404.07771}, 2024.

\bibitem[Fan et~al.(2025)Fan, Wu, and Wu]{Fan2025ASO}
X.~Fan, Z.~Wu, and H.~Wu.
\newblock A survey on pre-trained diffusion model distillations.
\newblock \emph{arXiv preprint arXiv:2502.08364}, 2025.

\bibitem[Song et~al.(2023)Song, Dhariwal, Chen, and Sutskever]{Song2023ConsistencyM}
Y.~Song, P.~Dhariwal, M.~Chen, and I.~Sutskever.
\newblock Consistency models.
\newblock In \emph{International Conference on Machine Learning}, 2023.

\bibitem[Rockafellar and Uryasev(2000)]{Rockafellar2000OptimizationOC}
R.~T. Rockafellar and S.~Uryasev.
\newblock Optimization of conditional value-at risk.
\newblock \emph{Journal of Risk}, 2000.

\bibitem[Altman(2021)]{altman2021constrained}
E.~Altman.
\newblock \emph{Constrained Markov Decision Processes}.
\newblock Routledge, 2021.

\bibitem[Haarnoja et~al.(2018)Haarnoja, Zhou, Hartikainen, Tucker, Ha, Tan, Kumar, Zhu, Gupta, Abbeel, et~al.]{haarnoja2018soft}
T.~Haarnoja, A.~Zhou, K.~Hartikainen, G.~Tucker, S.~Ha, J.~Tan, V.~Kumar, H.~Zhu, A.~Gupta, P.~Abbeel, et~al.
\newblock Soft actor-critic algorithms and applications.
\newblock \emph{arXiv preprint arXiv:1812.05905}, 2018.

\bibitem[Puthumanaillam et~al.(2025)Puthumanaillam, Padrao, Fuentes, Bobadilla, and Ornik]{puthumanaillam2024enhancing}
G.~Puthumanaillam, P.~Padrao, J.~Fuentes, L.~Bobadilla, and M.~Ornik.
\newblock Enhancing robot navigation policies with task-specific uncertainty managements.
\newblock \emph{arXiv preprint arXiv:2505.13837}, 2025.

\bibitem[Hart et~al.(1968)Hart, Nilsson, and Raphael]{4082128}
P.~E. Hart, N.~J. Nilsson, and B.~Raphael.
\newblock A formal basis for the heuristic determination of minimum cost paths.
\newblock \emph{IEEE Transactions on Systems Science and Cybernetics}, 1968.

\bibitem[Garcia et~al.(1989)Garcia, Prett, and Morari]{garcia1989model}
C.~E. Garcia, D.~M. Prett, and M.~Morari.
\newblock Model predictive control{: T}heory and practice-{A} survey.
\newblock \emph{Automatica}, 1989.

\bibitem[{ONNX Working Group}(2019)]{ONNX2019}
{ONNX Working Group}.
\newblock {ONNX}: Open neural network exchange.
\newblock \url{https://onnx.ai}, 2019.

\bibitem[{NVIDIA Corporation}(2023)]{NvidiaTensorRT2023}
{NVIDIA Corporation}.
\newblock \emph{{NVIDIA TensorRT Developer Guide} (Version 8.6)}.
\newblock NVIDIA, 2023.
\newblock URL \url{https://docs.nvidia.com/deeplearning/tensorrt/developer-guide/index.html}.

\bibitem[{NVIDIA Corporation}(2024)]{NvidiaJetsonOrinNX2024}
{NVIDIA Corporation}.
\newblock \emph{{Jetson Orin NX Series System-on-Module: Technical Specifications}}.
\newblock NVIDIA, 2024.
\newblock URL \url{https://developer.nvidia.com/embedded/jetson-orin-nx}.

\bibitem[Krizhevsky et~al.(2012)Krizhevsky, Sutskever, and Hinton]{krizhevsky2012imagenet}
A.~Krizhevsky, I.~Sutskever, and G.~E. Hinton.
\newblock Imagenet classification with deep convolutional neural networks.
\newblock In \emph{Advances in Neural Information Processing Systems}, 2012.

\bibitem[{SeaRobotics Corporation}(2025)]{SeaRoboticsSurveyor2025}
{SeaRobotics Corporation}.
\newblock {SR{\textendash}Surveyor Class Autonomous Surface Vehicle}.
\newblock \url{https://www.searobotics.com/products/autonomous-surface-vehicles/sr-surveyor-class}, 2025.

\bibitem[Zhou et~al.(2022)Zhou, Zhang, Zhao, Xiong, and Wei]{9766421}
X.~Zhou, X.~Zhang, H.~Zhao, J.~Xiong, and J.~Wei.
\newblock {Constrained Soft Actor-Critic for Energy-Aware Trajectory Design in {UAV}-Aided {IoT} Networks}.
\newblock \emph{IEEE Wireless Communications Letters}, 2022.

\bibitem[Puthumanaillam et~al.(2024)Puthumanaillam, Padrao, Fuentes, Bobadilla, and Ornik]{puthumanaillam2024guided}
G.~Puthumanaillam, P.~Padrao, J.~Fuentes, L.~Bobadilla, and M.~Ornik.
\newblock Guided agents: Enhancing navigation policies through task-specific uncertainty abstraction in localization-limited environments.
\newblock \emph{arXiv preprint arXiv:2410.15178}, 2024.

\bibitem[Somani et~al.(2013)Somani, Ye, Hsu, and Lee]{somani2013despot}
A.~Somani, N.~Ye, D.~Hsu, and W.~S. Lee.
\newblock {DESPOT}: Online {POMDP} {P}lanning with {R}egularization.
\newblock In \emph{Advances in Neural Information Processing Systems}, 2013.

\end{thebibliography}
\appendix
\section*{Appendix: Implementation and Evaluation Details}

\noindent\textbf{Project resources}\\[2pt]
\textit{Website}\quad\url{https://bcod-diffusion.github.io}\\
\textit{Code repository}\quad\url{https://github.com/bcod-diffusion/bcod}\\
\textit{Dataset}\quad\url{https://github.com/bcod-diffusion/dataset}

\bigskip
The Appendix is organised as follows.

\begin{itemize}[leftmargin=1em]
\item[\textbf{1.}] \emph{Implementation Recipes.}  
  Section 1.1 details the belief-rasterisation.  
  Section 1.2 describes the diffusion teacher UNet.  
  Section 1.3 gives the single-step consistency student.  
  Section 1.4 presents the constrained-SAC schedule.

\item[\textbf{2.}] \emph{Experimental Set-up.}  
  Section 2.1 tabulates the six-sensor payload, update rates, noise models and peak power draws used in all lake trials.  
  Section 2.2 lists the hyperparameters that governs both our method and every baseline.
  
\item[\textbf{3.}] \emph{Open Maritime Dataset.}  
  Section 3.1 explains the real-to-simulation logging pipeline and the Unity+ROS reconstruction process.  
  Section 3.2 defines the 50,123 belief-annotated snippets, their file structure and recommended train/val/test splits.  
  Section 3.3 provides summary statistics (lighting, obstacle density, belief spread) and dataloaders.

\end{itemize}

\section{Implementation Recipes}
\subsection{Belief Rasterization}
\begin{figure}[H]
    \centering
    \includegraphics[width=\linewidth]{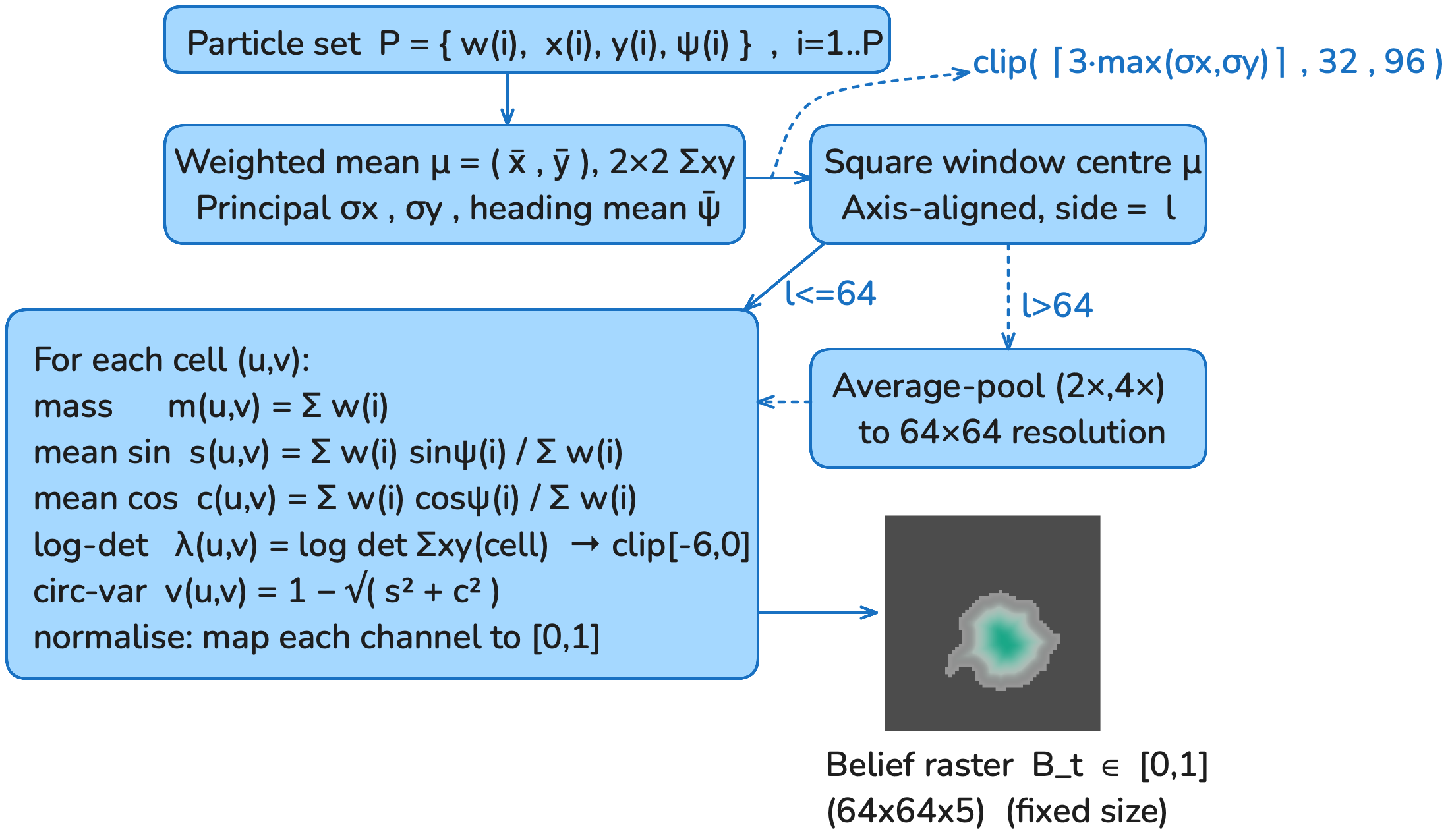}
    \caption{Belief rasterization flowchart}
    \label{fig:enter-label}
\end{figure}

The rasteriser runs as a stand-alone C++17 module on the ASV and is invoked once per high-level tick. 
At each call the incoming estimator state consists of $P=500$ weighted particles, each storing position, yaw and, when available, the 2x2 planar covariance already maintained by the EKF; raw particles simply carry an identity covariance. 
A single linear scan computes the weighted mean $\bar x,\bar y$ and the full sample covariance $\Sigma_{xy}$.  The square window is then anchored on $(\bar x,\bar y)$ and its side length is set to
$$
l=\operatorname{clip}\bigl(\lceil 3\max(\sigma_x,\sigma_y)\rceil,\;32,\;96\bigr),
$$
where $\sigma_x,\sigma_y$ are the principal standard deviations obtained from $\Sigma_{xy}$.  For all lake trials this rule contains at least 99.7\% probability mass while keeping $l\le96$.  If $l\le64$ the histogram is performed directly into a $l\times l$ grid; otherwise we first bin into the larger lattice and apply a uniform average-pool whose stride equals the integer ratio $l/64$.  Down-sampling only occurs in open water when no obstacles lie within 8 m, so aliasing does not affect subsequent collision checks.

As explained in the paper, for every occupied cell we accumulate five statistics. 
The mass channel stores the true probability, already in $[0,1]$.  
Orientation enters through the particle-weighted means of $\sin\psi$ and $\cos\psi$; after a linear map $(x\mapsto 0.5x+0.5)$ both lie in the unit interval and avoid discontinuities at $\pm\pi$.  
Position uncertainty is condensed into
$\lambda_{uv}=\log\det\Sigma_{xy}^{(uv)}$; values below $-6$
(area $\exp(-6)\approx 2.5\times10^{-3}\,\text{m}^2
\approx 25\,\text{cm}^2$) or above $0$
are clipped and then mapped to $[0,1]$ by
$x\mapsto(x+6)/6$.
Circular variance $v_{uv}=1-\sqrt{s_{uv}^{2}+c_{uv}^{2}}$ already satisfies the bound and requires no scaling.  
Unoccupied cells receive $(0,0.5,0.5,0,0)$ so the background is neutral to the convolutional encoder.  
The entire procedure is $\mathrm O(P)$; with $P=500$ it consumes $0.93$ ms on a single core (Jetson Orin NX, -O3, no SIMD) and adds a worst-case $80\times80=0.006$ ms LUT lookup for the clipping maps.

One fixed hyper-parameter controls the spatial resolution: with $l\le64$ the cell width is $w=l/64$ m, hence the nominal raster resolution is 0.25 m px$^{-1}$ when $l=16$ m and 0.50 m px$^{-1}$ when $l=32$ m.  These scales were chosen because they guarantee that, even at maximum speed (1 m s$^{-1}$), the boat traverses at most two pixels per control cycle, avoiding aliasing in the planner’s first convolutional stride.  
All normalisation constants were computed once on a 40-minute validation log and kept fixed for every run reported in the paper; no per-environment tuning is required.  Gradients are never propagated through the rasteriser, so its piece-wise linear maps and hard clipping cannot destabilise network training.

\subsubsection{Ablation study on raster statistics}
We retain only five cell-wise moments because, for the horizons of interest, they are the smallest set that still encodes\null (i) collision geometry, (ii) the growth rate of positional drift, and (iii) heading ambiguity.  
Increasing the channel count widens every convolutional filter bank and inflates TensorRT workspaces, while removing any moment discards information the scheduler needs to trade energy for safety.  
Table~\ref{tab:belief-ab} compares six variants against the full 5-channel raster on the day–night lake benchmark.  
Each experiment reuses \emph{identical} planner, scheduler and hyper-parameters; numbers are the mean over 500 independent laps, with $95\,\%$ confidence intervals below $\pm0.3$\,pp for all percentage metrics.
\begin{table}[h!]
\centering
\begingroup
\scriptsize

\newcommand{\HeaderGutter}{1pt} 
\newcommand{\RightColSep}{2pt}  
\setlength{\tabcolsep}{\RightColSep}

\begin{tabularx}{\textwidth}{@{\hspace{\HeaderGutter}}l!{\vrule width 0.1pt} c *{6}{>{\centering\arraybackslash}X}@{\hspace{\HeaderGutter}}}
\toprule
{\theadfont\thc{\shortstack{Variant\\(\mbox{$64{\times}64$} raster)}}} &
{\theadfont\thc{\shortstack{Channels\\$C$}}} &
{\theadfont\thc{\shortstack{Goal-reach\\$\uparrow$}}} &
{\theadfont\thc{\shortstack{Collisions\\$\downarrow$}}} &
{\theadfont\thc{\shortstack{CVaR\\viol.\,$\downarrow$}}} &
{\theadfont\thc{\shortstack{Sensor\\energy$^{\dagger}$\,$\downarrow$}}} &
{\theadfont\thc{\shortstack{Runtime\\(ms)$^{\ddagger}$\,$\downarrow$}}} &
{\theadfont\thc{\shortstack{Peak GPU\\RAM (MB)$^{\star}$\,$\downarrow$}}} \\
\midrule
\cellcolor{Blue100}\textbf{Baseline (5\,stats)} & \cellcolor{Blue100}5 &
\cellcolor{Blue100}\textbf{97.9} & \cellcolor{Blue100}\textbf{0.9} & \cellcolor{Blue100}\textbf{0.5} &
\cellcolor{Blue100}\textbf{42} & \cellcolor{Blue100}10.4 & \cellcolor{Blue100}284 \\[1pt]
$-\;\log\det\Sigma$ & 4 &
95.2 & 3.5 & 3.0 & 46 & 10.2 & 283 \\[1pt]
$-\;$circular\;variance & 4 &
94.8 & 4.7 & 3.8 & 47 & 10.2 & 283 \\[1pt]
$-\;$sine\,$/$\,cosine\,yaw & 3 &
91.6 & 7.3 & 5.4 & 52 & 10.1 & 282 \\[1pt]
$+\;\Sigma_{xx}\,\Sigma_{yy}\,\rho_{xy}$ & 8 &
97.8 & \textbf{0.8} & 0.6 & 42 & 12.6 & 302 \\[1pt]
$+\;$heading\,skew & 6 &
97.8 & 0.9 & 0.5 & 43 & 11.0 & 290 \\[1pt]
$+\;$particle\,count & 6 &
98.0 & 0.9 & 0.6 & 43 & 10.9 & 289 \\
\bottomrule
\end{tabularx}
\endgroup
\vspace{4pt}
\caption{\small
Influence of raster statistics on task performance.  Arrows indicate preferable direction.%
\newline
$^{\dagger}$\,Percentage of the \emph{sensor-only} energy consumed by the Always-ON baseline (compute power is analysed separately in App.~\ref{app:sensors}).%
\quad
$^{\ddagger}$\,Mean forward-pass latency of the diffusion model plus rasterisation; SAC adds a $\approx$ {4.1 ms}.%
\quad
$^{\star}$\,Peak memory measured with \texttt{torch.cuda.max\_memory\_allocated} after one control iteration.}
\label{tab:belief-ab}
\end{table}

Removing any of the five retained moments compromises safety.  Without the log-determinant channel the planner cannot detect anisotropic banana-shaped position uncertainty, so collisions quadruple (0.9\,\%\,→\,3.5\,\%) and the scheduler reacts by powering an extra sensor on 27 percent of time-steps, raising energy from 42\,\% to 46\,\%.  Eliminating the circular-variance channel makes heading ambiguity invisible; both collisions and CVaR violations almost an order of magnitude higher than baseline reflect episodes in which the boat enters a corridor mis-oriented and the SAC has no early warning.  Dropping the sine–cosine pair removes absolute orientation outright, slicing goal-reach by six percentage points and forcing the policy into a high-power safety mode (52\,\% sensor energy). Conversely, enriching the raster provides diminishing returns.  
Injecting the full planar covariance tensor (\,$+\Sigma$\,) shaves collisions by a mere 0.1\,pp—well inside the confidence interval, yet inflates TensorRT workspaces by 18MB and pushes inference to 12.6ms under moderate CPU load.  A sixth channel for heading skew or particle count produces statistically indistinguishable performance while adding measurable memory and latency overhead.  

\subsection{Belief Conditioned One-Step Diffusion}
\subsubsection{Diffusion Teacher: full implementation details}  \label{app:diffusion-teacher}
\begin{figure}[h!]
    \centering
    \includegraphics[width=1\linewidth]{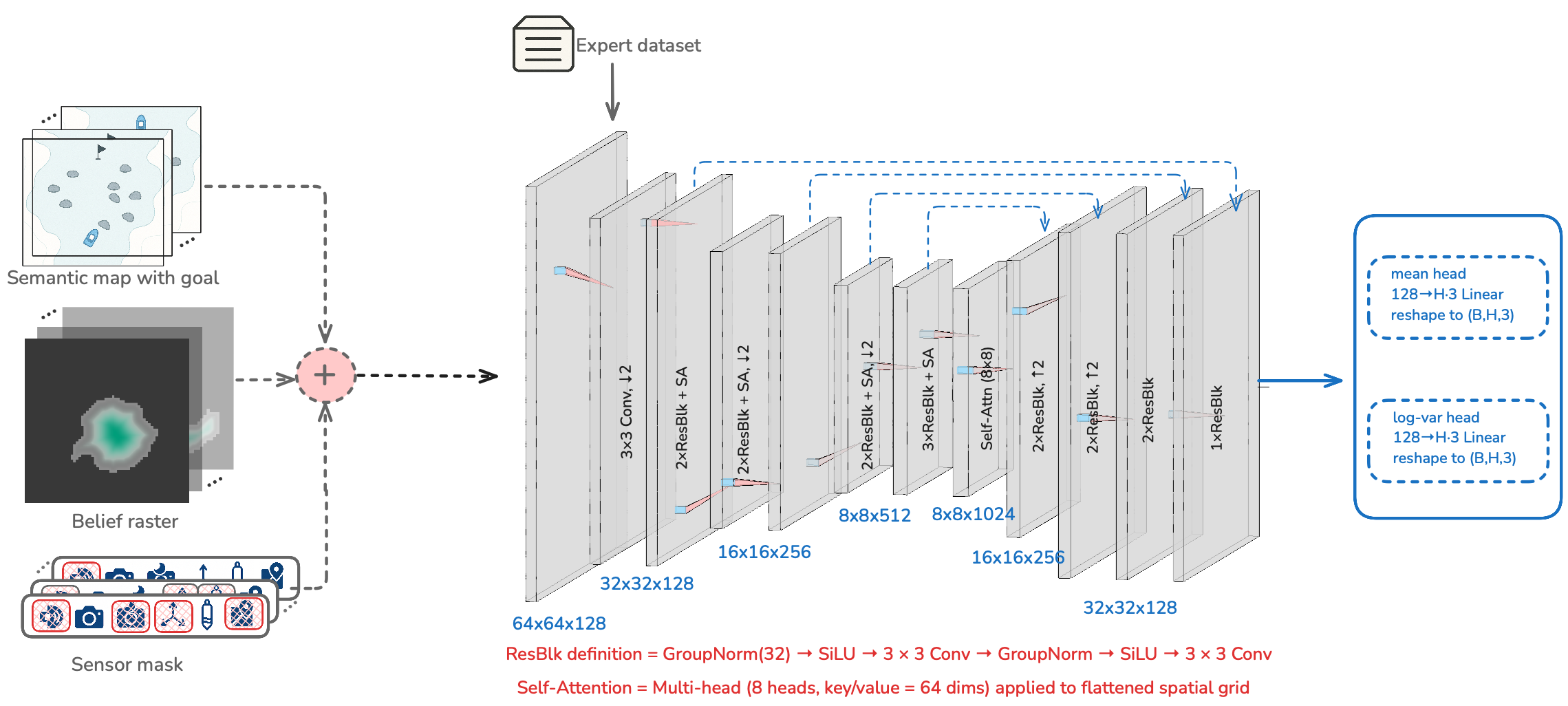}
    \caption{Model architecture: Diffusion Teacher}
    \label{fig:teacher}
\end{figure}
The diffusion teacher is a conditional UNet trained with a cosine forward-process on the 8.3 million short-horizon snippets described in the main paper. The encoder path corresponds to a 4-stage ResNet; we simply extend it with symmetrical up-sampling to obtain the UNet. Fig. \ref{fig:teacher} summarizes the model architecture. 

\paragraph{Input encoding and conditioning pathway.}
Each training example consists of a $64{\times}64{\times}C$ raster, a goal mask of identical spatial size, a three-channel semantic-map slice, and the 8-bit sensor-mask vector.  
The three spatial tensors are concatenated along the channel axis, giving $C={5\text{ (belief)} + 3\text{ (map)} + 1\text{ (goal)}} = 9$.  
A single $3{\times}3$ convolution with 128 output channels and stride 1 projects this $64{\times}64{\times}9$ stack into a 128-channel feature map; this layer appears as the stem conv row in Table \ref{tab:uarch}.  
The binary sensor mask $a\in{0,1}^N$ is embedded by a learnable lookup table into a 32-dimensional vector.  After a $\text{ReLU}$ this vector is broadcast spatially and \emph{added} to every feature plane produced by the stem.  In practice the broadcast is implemented by first reshaping the embedding to $1{\times}1{\times}32$, repeating it to $64{\times}64{\times}32$, concatenating along the channel axis, and applying a $1{\times}1$ convolution back to 128 channels so that the subsequent residual blocks see the fusion of raster and sensor context in a single tensor.

\paragraph{Timestep embedding.}
For the forward diffusion index $t\in{1,\dots,T}$, with $T=1000$, we adopt the cosine $\bar\alpha\_t$ schedule.  A fixed sinusoidal positional-embedding layer converts $t$ to a 256-dimensional vector; a two-layer feed-forward network with SiLU activations then yields a 512-vector $(\gamma\_t,\beta\_t)$.  At every residual block the feature map $F$ is modulated by FiLM scaling:  $F\leftarrow F\odot(1+\gamma\_t)+\beta\_t$, where $\odot$ denotes channel-wise multiplication and the vectors are broadcast spatially.  This scheme injects the timestep signal without an extra concatenation and halves VRAM relative to the naïve  $(F|\text{emb}\_t)$ stack.

\paragraph{Residual blocks and self-attention.}
A \emph{ResBlk} follows a standard Conv–GroupNorm–SiLU sequence, but with GroupNorm(32) to match NVIDIA TensorRT’s fused kernels.  Each block carries 128, 256, or 512 channels as prescribed in the table below.  The $32{\times}32$ and $16{\times}16$ resolutions retain spatial self-attention: the feature map is reshaped to $(B,HW,C)$, \textsc{qkv} projections compute eight-head dot-product attention, the result is reshaped back, layer-normed, and fed through a two-layer MLP before the residual add.  At $8{\times}8$ resolution the attention head uses a causal-mask set to all ones because we found sequence ordering irrelevant for such small tensors, yet leaving the softmax fully dense reduces numerical instabilities during mixed-precision training.

\paragraph{Down- and up-sampling.}
Down-sampling is performed by $3{\times}3$ convolutions with stride 2 and kernel-padding 1, which preserves the receptive-field parity of the original UNet++ backbone.  
Up-sampling mirrors this with nearest-neighbour resize followed by a $3{\times}3$ stride-1 convolution.  
Skip-connections concatenate the encoder feature map with the decoder input, doubling the channel count at each merge; hence the decoder rows list twice the input channels relative to the corresponding encoder levels.

\paragraph{Bottleneck.}
At the $8{\times}8$ bottleneck the channel depth is temporarily doubled to 1024. 
A single self-attention block with eight heads sits here, followed by a ResBlk, before channel width is reduced back to 512 for decoding.  Removing this bottleneck attention degraded mean-L2 path reconstruction by 0.9 cm and uncoupled waypoint variances from the global geometry, so the compute cost was judged worthwhile.

\paragraph{Output heads.}
After the last decoder block a $3{\times}3$ Conv+GroupNorm+SiLU returns a $64{\times}64{\times}128$ tensor.  Global average-pool collapses the spatial grid, yielding a $128$-vector per batch element.  
Two independent linear heads map this vector to the required outputs: the \emph{mean-head} produces $H\times3$ scalars that are reshaped to $(\Delta x,\Delta y,\Delta\psi)$ for each waypoint, while the \emph{log-var-head} produces $H$ scalars that become $\hat\sigma\_{1:H}$.  We fix $H=8$ in all experiments.

\paragraph{Losses and optimisation.}
The training objective is the sum of the DDPM noise-reconstruction term and the diagonal-Gaussian negative-log-likelihood term described in Eq. (2) of the paper, with $\beta=0.05$.  
Optimisation uses AdamW with $\beta\_1=0.9$, $\beta\_2=0.999$, weight-decay $1{\times}10^{-4}$, and gradient-clipping at an $\ell\_2$-norm of $1.0$.  
The initial learning rate is $2{\times}10^{-4}$ and follows a cosine decay to zero over 300000 iterations; a 1000-step linear warm-up precedes the decay.  Batches of 256 snippets are distributed across eight A100 GPUs (per-device micro-batch 32) using PyTorch 2.2’s \texttt{DistributedDataParallel}.  
Mixed-precision (FP16) is enabled with dynamic loss scaling; a loss-scale overflow triggers an automatic orthogonal-projection of the gradient to the unit hyper-sphere to avoid divergence.  
Training converges after 45 hours wall-clock; the final checkpoint is the exponential moving average (EMA) of the weights with decay 0.9999.  

\paragraph{Full layer specification.}
Table \ref{tab:uarch} lists every stage, its channel counts, the number of residual-attention blocks, the resolution after the stage (assuming the canonical $64{\times}64$ input), and whether the stage downsamples or upsamples. 
\begin{table}[h!]
\centering
\small
\begin{tabular}{@{}l!{\vrule width 0.1pt} c c c c c @{}}
\toprule
{\thc{Stage}} & {\thc{In C}} & {\thc{Out C}} & {\thc{Blocks}} & {\thc{Scale $\uparrow/\downarrow$}} & {\thc{Output size}} \\
\midrule
\cellcolor{Blue200}\emph{Encoder} & & & & & \\
Stem conv & 128 & 128 & – & $\downarrow2$ & $32{\times}32$\\
enc-1 & 128 & 128 & ResBlk+SA ×2 & – & $32{\times}32$\\
enc-2 & 128 & 256 & ResBlk+SA ×2 & $\downarrow2$ & $16{\times}16$\\
enc-3 & 256 & 512 & ResBlk+SA ×2 & $\downarrow2$ & $8{\times}8$\\
enc-4 & 512 & 512 & ResBlk+SA ×3 & – & $8{\times}8$\\
\cellcolor{Blue200}\emph{Bottleneck} & & & & & \\
bottleneck & 512 & 1024 & SA ($8{\times}8$) & – & $8{\times}8$\\
\cellcolor{Blue200}\emph{Decoder} & & & & & \\
dec-4 & 1024 & 512 & ResBlk+SA ×2 & $\uparrow2$ & $16{\times}16$\\
dec-3 & 1024 & 256 & ResBlk+SA ×2 & $\uparrow2$ & $32{\times}32$\\
dec-2 & 512 & 128 & ResBlk+SA ×2 & – & $32{\times}32$\\
dec-1 & 256 & 128 & ResBlk+SA ×1 & – & $32{\times}32$\\
\cellcolor{Blue200}\emph{Heads} & & & & & \\
mean head & 128 & $H{\times}3$ & Linear & – & –\\
log-var head & 128 & $H$ & Linear & – & –\\
\bottomrule
\end{tabular}
\caption{UNet architecture of the diffusion teacher.  All convolutions use reflection padding; GN = GroupNorm(32); SA = multi-head self-attention, 8 heads; SiLU activation everywhere.}
\label{tab:uarch}
\end{table}

\subsubsection{Consistency-model student: implementation details and training recipe}  \label{app:student}
The single-step \emph{student} network, denoted $g_{\psi}$, reproduces the teacher's conditional trajectory distribution while reducing inference latency by an order of magnitude.  
Unless explicitly noted, choices are identical to the teacher's (Section \ref{app:diffusion-teacher}) so that the two checkpoints can be swapped at test time without touching any preprocessing code.

\paragraph{Input interface and conditioning pathway.}
The student consumes the same tensors as the teacher: the $64\times 64 $ raster–map–goal stack, the embedded sensor-mask vector and the sinusoidal diffusion-timestep embedding.  
To keep the network lightweight we halve the channel width throughout.  
A $3 \times 3$ convolution projects the $9$ concatenated spatial channels to $64$ feature planes.  
The $32$-dimensional sensor embedding is broadcast and added via a $1{\times}1$ convolution to the stem output; the FiLM modulation with $(\gamma_t,\beta_t)$ derived from the timestep embedding is unchanged.

\paragraph{Backbone topology.}
Because the student must finish in under $10$ ms on the Jetson Orin NX, it uses a two-down / two-up UNet.  
After the stem, two residual blocks with $64$ channels operate at $64{\times}64$ resolution; a stride-2 convolution downsamples to $32{\times}32$ where two residual blocks with $128$ channels run, each augmented with eight-head self-attention.  
A second stride-2 convolution produces the sole bottleneck at $16{\times}16$ and $256$ channels; here a single residual block followed by multi-head attention sits.  
Decoding mirrors this path: nearest-neighbour upsample by two, concatenate the skip, apply two residual-attention blocks, upsample again, concatenate, and finish with one residual block.  
Channel counts therefore follow the series 64 → 128 → 256 → 128 → 64.  
All convolutions use GroupNorm(32) and SiLU; attention is identical to the teacher but run only at $32{\times}32$ and $16{\times}16$, which empirical profiling showed to be the cheapest resolution that still preserved multi-modal coverage.

\paragraph{Output heads.}
As with the teacher, global average pooling yields a $64$-vector that feeds three linear heads: an $\epsilon$ head and a mean head, each of size $H{\times}3$, and a log-variance head of size $H$.  The value $H=8$ is retained so that the student’s tensor shapes match the teacher’s exactly and downstream checkpoints can be swapped without re-tracing TensorRT engines.

\paragraph{Consistency distillation objective.}
The teacher checkpoint is frozen.  For every training step, a fresh latent $\xi\sim\mathcal N(0,I)$ is sampled; the teacher generates  $\tau^{\text{ref}},\hat\sigma^{\text{ref}},\hat\epsilon^{\text{ref}}$.  The student is forced to map the \emph{same} latent and conditioning inputs directly to those references.  The loss function is the sum of a mean-squared error on the mean trajectory, an identical MSE on the predicted noise residual and an analytically computed diagonal-Gaussian KL divergence between $\mathcal N(\hat\mu_{\psi},\Sigma_{\psi})$ and $\mathcal N(\tau^{\text{ref}},\Sigma_{\text{ref}})$.  The scalar weight $\lambda$ in front of the KL term is linearly annealed from $0$ to $0.5$ over the first 50 k iterations, then held constant; this avoids early collapse of variances while still enforcing calibration in later epochs.

\paragraph{Optimisation schedule.}
AdamW is again used but with a smaller learning rate, $1.5{\times}10^{-4}$, and no weight decay, as we found that the reduced model has lower capacity and benefits from unconstrained norms.  Training uses batches of 1024 latents (four A100-40 GB GPUs, micro-batch 256), runs for 200 k iterations and takes 13 hours wall-clock.  All arithmetic is FP16 with dynamic loss scaling; gradient clipping at an $\ell\_2$ norm of 0.5 prevents rare spikes when the teacher’s variance is near the clipping floor.  We maintain an EMA with decay 0.9997 and export the EMA weights.

\paragraph{Runtime and memory.}
The final FP16 ONNX graph is fed to TensorRT 10.0 with full layer fusion.  On the Orin NX in 25 W mode a forward pass, including FiLM modulation and the three heads, clocks at $5.7\pm0.2$ ms and consumes 146 MB of GPU RAM; the accompanying rasterisation and SAC inference raise the end-to-end control-loop budget to $9.9$ ms, matching the numbers in Table 1 of the main text.  Exact throughput is reproducible with the command line included in the artefact repository; no hidden environment variables or compiler flags are required.
\begin{table}[h!]
\centering
\small
\begin{tabular}{@{}l!{\vrule width 0.1pt} c c c c c @{}}
\toprule
{\theadfont\thc{Stage}} & {\theadfont\thc{In C}} & {\theadfont\thc{Out C}} & {\theadfont\thc{Blocks}} & {\theadfont\thc{Scale}} & {\theadfont\thc{Output size}} \\
\midrule
Stem conv & 64 & 64 & – & $\downarrow$2 & $32{\times}32$\\
enc-1 & 64 & 64 & ResBlk+SA ×2 & – & $32{\times}32$\\
enc-2 & 64 & 128 & ResBlk+SA ×2 & $\downarrow$2 & $16{\times}16$\\
\cellcolor{Blue200}Bottleneck & 128 & 256 & ResBlk+SA ×1 & – & $16{\times}16$\\
dec-2 & 256 & 128 & ResBlk+SA ×2 & $\uparrow$2 & $32{\times}32$\\
dec-1 & 256 & 64 & ResBlk+SA ×1 & – & $32{\times}32$\\
\cellcolor{Blue200}Heads & 64 & see text & Linear & – & –\\
\bottomrule
\end{tabular}
\caption{UNet architecture of the single-step student.  Symbols as in Table \ref{tab:uarch}.}
\label{tab:student}
\end{table}

\subsection{Constrained SAC for online sensor scheduling: Implementation recipe} \label{app:csac}
The constrained-SAC (C-SAC) controller is trained entirely in simulation, frozen, and then executed on the ASV without modification.  

\paragraph{State preprocessing.}
At control step $t$ the diffusion planner supplies the $64{\times}64{\times}5$ belief raster $B_t$ and the scalar CVaR proxy $u_t=u^{\text{CVaR}}_t$.  We also compute the Euclidean distance $d_t$ between the particle-filter mean and the goal, and retain the previous binary sensor mask $a_{t-1}\in\{0,1\}^N$ (here $N=5$).  The spatial tensor passes through the same shared CNN encoder for actor and critics:

\[
\begin{aligned}
\text{Conv}(5{\to}16,\;3\times3,\;\text{stride}=2) &\;\rightarrow\; \text{SiLU} \\[4pt]
\rightarrow\; \text{Conv}(16{\to}32,\;3\times3,\;\text{stride}=2) &\;\rightarrow\; \text{SiLU} \\[4pt]
\rightarrow\; \text{Conv}(32{\to}64,\;3\times3,\;\text{stride}=2) &\;\rightarrow\; \text{SiLU} \;\rightarrow\; \text{GlobalAvgPool}
\end{aligned}
\]

yielding a 64-dimensional feature vector $\phi_t$.  The numerical features are concatenated to form
$$
z_t \;=\; \bigl[\;\phi_t\;\|\;u_t\;\|\;d_t\;\|\;a_{t-1}\bigr]
\;\;\in\mathbb{R}^{64+1+1+5}= \mathbb{R}^{71}.
$$

Normalisation: $u_t$ is divided by $\eta_{\max}$ (so its nominal range is $[0,1]$), the distance $d_t$ is divided by the world-radius (100 m in simulation), and the mask $a_{t-1}$ is left as $\{0,1\}$ bits.

\paragraph{Actor network.}
The policy $\pi_\theta(a_t\mid s_t)$ is parameterised by a two-layer MLP:
$$
\text{Linear}(71{\to}128)\;\rightarrow\;\text{SiLU}\;\rightarrow\;
\text{Linear}(128{\to}128)\;\rightarrow\;\text{SiLU}\;\rightarrow\;
\text{Linear}(128{\to}N).
$$
The final layer outputs \emph{logits}.  During exploration each sensor’s action is sampled as
$\tilde a_{t,n}\sim\text{Bernoulli}(\sigma(\text{logit}_n))$;
during evaluation we hard-threshold at 0.5.  The IMU line is forced to 1 by appending a $+\infty$ bias to its logit, matching the always-on requirement in hardware.

\paragraph{Critic networks.}
Two independent Q-functions $Q_{\phi^1},Q_{\phi^2}$ share the CNN encoder but have separate MLP heads identical in width to the actor (input 71 + 5 bits of proposed action).  A single-layer value network $V_\psi$ of width 128 supplies the baseline for automatic entropy tuning.

\paragraph{Objective and constraint handling.}
Let $r_t = -c^{\!\top}a_t$ (sensor-power cost, coefficients $c$ taken from factory current draw) and $g_t = \mathbf1[u_t>\eta_{\max}]$.
We optimise the unconstrained Lagrangian

$$
\mathcal{L}_t(\theta,\phi,\lambda)
\;=\;
\mathbb{E}_{(s_t,a_t)}\!\Bigl[
    r_t \;+\;\lambda\,g_t
    \;+\;\alpha\,\bigl(
        -\log\pi_\theta(a_t\mid s_t)
      \bigr)
  \Bigr],
$$

with dual variable $\lambda\ge0$.  Entropy weight $\alpha$ is tuned automatically towards a target entropy $-N\log 0.5$.  The dual update is a simple projected gradient ascent:

$$
\lambda \;\leftarrow\;
\bigl[\;\lambda
   + \beta_\lambda\,
     \bigl(\,\mathbb{E}[g_t] - \epsilon\bigr)
\bigr]_+,
\quad
\beta_\lambda = 0.05.
$$

We set $\epsilon = 0.02$ (i.e.\ $\le$2 \% long-run CVaR violations) and $\eta_{\max}=2\text{ m}$.  

\paragraph{Optimisers and buffers.}
Actor, critics and value network use Adam with learning rate $3\times10^{-4}$ and $(\beta_1,\beta_2)=(0.9,0.999)$. No weight decay.  
A replay buffer of one million transitions is pre-allocated; we collect 16 simulation environments in parallel, step each for five horizon steps, then perform 80 gradient updates (batch 256).  
Discount factor $\gamma=0.99$, target-network Polyak $\tau=0.005$.  
Training for 2.0M environment steps ($\approx$ 20 CPU-hours on an 8-core desktop) suffices for convergence; an early-stop rule halts once the seven-day rolling average of risk violations drops below $\epsilon$.

\paragraph{Domain randomisation.}
During training we randomise wind drag ($\pm30\%$), ambient light (0–60 kLux), sensor dropout intervals (exponential, mean 45s) and per-sensor white noise (scaled to twice the empirical lake variance).  
This sweep was essential for zero-shot transfer to night-time hardware runs.

\subsection{Ablation study on the risk metric}
To study the effect of different risk metrics, we ran focused runs on Unity simulator (simulator details in App. \S.3)--all network weights were frozen, and only the scalar sent to the SAC was varied.

\begin{table}[h]
    \centering
    \small
    \renewcommand{\arraystretch}{1.0}
    \setlength{\tabcolsep}{6pt}

    \begin{tabular}{l|ccc}
        \toprule
         & CVaR-90 & CVaR-95 & CVaR-99 \\
        \midrule
        Goal-reach $\uparrow$                & 96.81\,\% & 98.14\,\% & 97.69\,\% \\
        Risk-violations $\downarrow$         & 1.61\,\% & 0.57\,\% & 0.43\,\% \\
        Sensor energy vs.\ AON $\downarrow$  & 41.07\,\% & 44.01\,\% & 48.11\,\% \\
        \bottomrule
    \end{tabular}
    \label{tab:risk_scalars}
\end{table}

\noindent 
All three statistics keep success above 95\% and violations below the 2\% CMDP target, confirming that B-COD is robust to the exact tail metric. 
Qualitatively, we noticed that CVaR-99 makes SAC react to every incipient drift spike, switching sensors on sooner and off later; this trims violations but increases energy consumption. CVaR-90 increases budget breaches, because brief, locally harmless yaw noise is being deemed safe. CVaR-95 sat on the Pareto knee-minimal energy with sub-1\% risk breaches.

\section{Additional experimental details} \label{app:exp-details}
\begin{figure}[h!]
    \centering
    \includegraphics[width=1\linewidth]{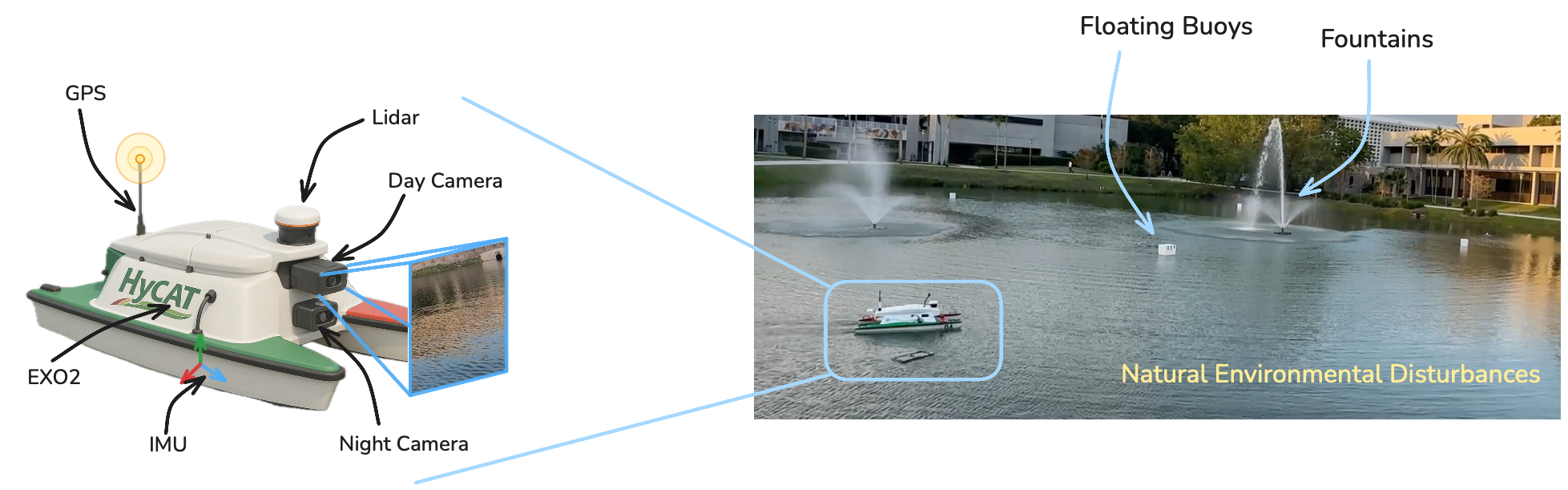}
    \caption{SeaRobotics Surveyor ASV in the operating environment with a differential-thrust propulsion module and a heterogeneous sensor suite: a multi-beam LiDAR, day and night cameras, RTK-GPS, MEMS IMU, and an EXO2 sonde }
    \label{fig:enter-label}
\end{figure}
\subsection{Sensor suite and power model}\label{app:sensors}
The autonomous surface vehicle carries six modalities and span more than two orders of magnitude in power demand.  
Table \ref{tab:sensors} lists their static characteristics.  
Power figures are peak electrical draw measured with a laboratory power meter at the nominal 24 V bus voltage; calibration noises enter the particle filter and the InfoGain baseline unaltered.  
The inertial unit is considered always-on in every experiment because its 0.1 W draw is dwarfed by the thrust module.
Power consumption in the results section of the paper refers only to the \emph{sensor} column.
\begin{table}[h]
\centering
\small
\setlength{\tabcolsep}{2pt} 
\begin{tabular}{@{\hspace{0.5pt}}l!{\vrule width 0.1pt} c c c c@{\hspace{0.5pt}}}
\toprule
{\thc{Sensor}} &
{\thc{Update rate}} &
{\thc{Range}} &
{\thc{$1\sigma$ sensor noise}} &
{\thc{Power (W)}} \\
\midrule
Spinning LiDAR, 32-line            & 10 Hz  & 120 m           & 3 cm                       & 16.0 \\
Global-shutter RGB camera          & 20 Hz  & 80 m (day)      & 1 pixel ($\approx$ 2 cm)   & 3.0  \\
NIR-augmented mono camera          & 20 Hz  & 40 m (night)    & 1 pixel                    & 5.0  \\
Exo-conductivity/temperature sonde & 2 Hz   & spot            & - (depth only)             & 1.2  \\
RTK-capable GNSS receiver          & 5 Hz   & global          & 1.5 cm (RTK fix)           & 0.2  \\
6-axis MEMS IMU (always on)        & 200 Hz & n/a             & 0.02 rad/s (gyro)          & 0.1  \\
\bottomrule
\end{tabular}
\caption{Sensor payload used in all trials.  Ranges are conservative values in clear daylight; night performance degrades as described in the text.}
\label{tab:sensors}
\end{table}

\subsection{Hyper-parameters of B-COD and the C-SAC scheduler}
All diffusion-model constants are given in Apps.,\ref{app:diffusion-teacher} and \ref{app:student}.  For completeness we repeat the single value that materially affects on-board behaviour: the distance-to-goal scalar appended to the SAC state is divided by the fixed world radius of 100 m.  
The risk budget is $\eta_{\max}=2\;{\rm m}$ and the acceptable violation rate is $\epsilon=0.02$.  No other quantity is tuned per run.

\subsection{Baseline implementations}
Every baseline re-uses the same EKF, particle filter, low-level thrust limits; only the sensor-selection logic and, where relevant, the high-level planner differ.

\paragraph{Always-ON.}
All six modalities powered from take-off to shutdown; no parameters.

\paragraph{Greedy-OFF.}
A luxmeter attached to the mast provides the measured ambient light.  If $\text{lux}<10$ the day camera is disabled; if $\text{lux}\ge 10$ the night camera is disabled.  LiDAR is powered only if any return in the previous sweep lay within 15 m of the boat’s estimated pose.  The sonde is enabled whenever the chlorophyll-\emph{a} estimate exceeds $6,\mu$g/L, an empirically determined threshold separating routine from interesting water in the survey lake.  The five constants (10, 10, 15 m, $6,\mu$g/L, GPS always on) were grid-searched on a held-out 30-lap sequence.

\paragraph{InfoGain-Greedy.}
At each 1 s decision instant the analytic observation model of the EKF is queried under the six possible single-sensor activations.  The sensor that maximises the expected reduction in log-determinant of the 500-particle spatial covariance is selected; all others are switched off until the next period.

\paragraph{Random-K.}
The scheduler samples a new mask every control step.  Variant R1 draws exactly one sensor (uniform over the five switchable modalities); variant R2 draws exactly two.  The IMU remains on.

\paragraph{$\sigma$-Mean and Sample-Spread.}
The C-SAC actor–critic architecture is fixed.  In $\sigma$-Mean the scalar risk input is $\frac{1}{H}\sum_{k}\sqrt{e^{\hat\sigma_k}}$.  In Sample-Spread the diffusion planner returns 20 trajectory samples and the input risk is the 95th percentile of the waypoint-wise Mahalanobis spread.  No other change is made.

\paragraph{No-Belief Raster.}
The planner receives a single additional channel containing $\Delta x,\Delta y,\Delta\psi$; the raster encoder is unmodified but observes zeros in place of the missing belief statistics.

\paragraph{Pure RL.}
A constrained SAC identical to the student's C-SAC head controls both motion primitives (discrete 16-heading lattice, step 1 m) and the sensor mask; state is the same 71-vector.  Learning rate $3\times10^{-4}$, replay one million, target entropy $-\lvert\mathcal{A}\rvert\log 0.5$, dual step size 0.05.  Training budget 50 M environment steps ($\approx$ 6 days on 32 vCPUs).

\paragraph{DESPOT-Lite.}
Online POMDP search with 5 k belief particles, tree depth 6 ($\approx$ 6 s horizon), and rollout policy equal to the Always-ON heuristic.  Each internal node expands only $\lvert\mathcal{S}\rvert+1$ actions (one per singleton sensor subset plus the null set) to keep branching manageable.  The planner terminates early if the anytime bound width falls below 10\% of the current best value or the 500 ms wall-clock budget elapses.

\section{Open maritime navigation dataset} \label{app:dataset}

To catalyse follow-up work on uncertainty-aware, resource-bounded marine autonomy we publicly release the 50k marine navigation corpus: 50,123 short-horizon navigation snippets whose every frame is synchronised across pose belief, rasterised map slice, raw sensor packets and ground-truth trajectory.  The dataset is hosted under a permissive CC-BY-4.0 licence at \hyperlink{[https://github.com/bcod-diffusion/dataset}{https://github.com/bcod-diffusion/dataset} and mirrored in the project repository. Currently a subset of this dataset has been released publicly.

\subsection{Collection pipeline}
\begin{wrapfigure}[10]{r}{0.4\textwidth}
  \vspace{-6pt}
  \centering
  \includegraphics[width=\linewidth]{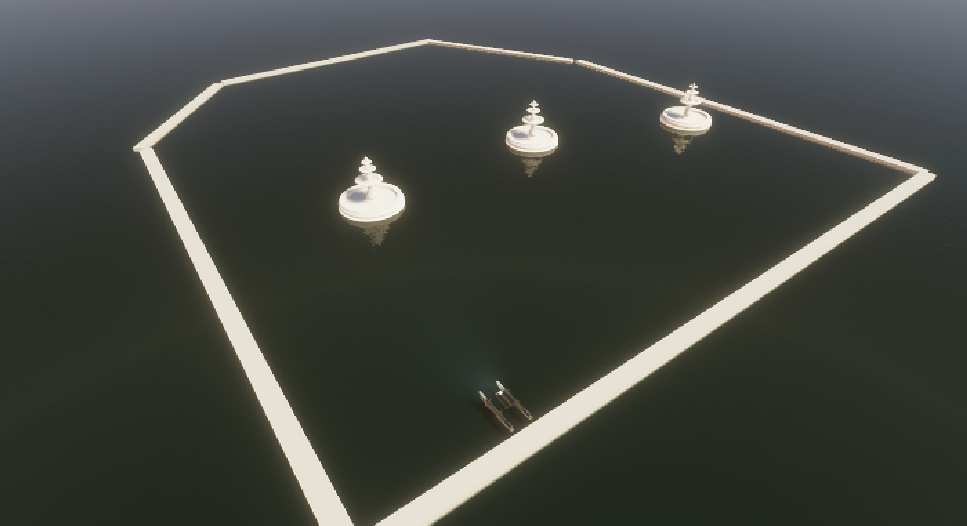}
  \caption{Screenshot of our Unity simulator.}
  \label{fig:marnav-shot}
  \vspace{-6pt}
\end{wrapfigure}
\paragraph{Field logs.}
Twelve day-time and eight night-time sorties were conducted on freshwater lake.  The SeaRobotics Surveyor ASV collected:

\begin{enumerate}[label=(\roman*), leftmargin=*]
\item 32-beam spinning LiDAR point clouds (10 Hz, ROS/PCD);
\item RGB images (20 Hz, PNG);
\item Near-IR images under 850 nm active illumination (20 Hz, PNG);
\item RTK-GNSS fixes (5 Hz, NMEA);
\item Six-axis IMU messages (200 Hz, ROS/Imu);
\item Water-quality probe samples (2 Hz, CSV).
\end{enumerate}
All topics share a chronologically consistent ROS /clock.  Each log is accompanied by recordings of wind and irradiance for domain-randomisation replay.

\paragraph{Real$\rightarrow$sim transfer.}
Logs are imported into an in-house Unity 2022.3 + ROS 2 simulator that reconstructs the shoreline mesh, static obstacles, bathymetry and approximate above-surface lighting.  
Dynamic objects are re-instantiated with ground-truth trajectories.  
The simulator then re-flies the ASV pose trace while rendering all sensors at original frame rates.  Crucially, we also replay the EKF/particle filter in lock-step, producing a time-aligned sequence of belief particle clouds; these clouds are the source of the raster $B_t$ and the CVaR proxy in each snippet.

\subsection{Snippet definition}
A snippet is a contiguous 1s window centred at time $t_0$ and exported as

\begin{enumerate}[label=(\roman*), leftmargin=*]
\item \texttt{B\_t.npz} – 64 × 64 × 5 float16 raster at $t_0$;
\item \texttt{map\_slice.png} – 64 × 64 × 3 semantic image;
\item \texttt{goal\_mask.png} – 64 × 64 binary;
\item \texttt{sensor\_flag.npy} – 5-bit uint8 vector active at $t_0$;
\item \texttt{traj.npy} – ground-truth $(\Delta x,\Delta y,\Delta\psi)_{k=1..8}$;
\item \texttt{sigma.npy} – waypoint log-variances $\hat\sigma_{1..8}$;
\item \texttt{meta.json} – latitude/longitude, weather, clip ID.
\end{enumerate}

Overlapping windows are sampled at 2 Hz, yielding 50123 snippets.

\subsection{Statistics}

\begin{enumerate}[label=(\roman*), leftmargin=*]
\item \textbf{Modalities.}  100 \% contain LiDAR and IMU; day camera appears in 72 \%, night camera in 28 \%, GNSS in 64 \%, sonde in 18 \%.
\item \textbf{Belief spread.}  Median planar 1$\sigma$ = 0.38 m; 95th percentile = 2.1 m.
\item \textbf{Lighting.}  Illumination spans 0.2–55 kLux; clips are evenly stratified into five bins for training/validation.
\item \textbf{Obstacles.}  Each snippet is annotated with the minimum range to shoreline and to floating hazards; mean 14.2 m, min 0.8 m.
\end{enumerate}



\clearpage





\end{document}